\documentclass[lettersize,journal]{IEEEtran}
\usepackage{amsmath,amsfonts}
\usepackage{algorithmic}
\usepackage{algorithm}
\usepackage{array}
\usepackage{subfigure}
\usepackage{textcomp}
\usepackage{stfloats}
\usepackage{url}
\usepackage{verbatim}
\usepackage{graphicx}
\usepackage{cite}
\usepackage{enumitem}
\usepackage{amsmath}
\usepackage{amssymb}
\usepackage{empheq}
\usepackage{tabularx}
\usepackage{hyperref}
\usepackage{stfloats} 

\begin{document}

\title{Multi-Scale Deformable Transformers for Student Learning Behavior Detection in Smart Classroom}

\author{Zhifeng Wang, Minghui Wang, Chunyan Zeng, Longlong Li

\thanks{This work was supported by the Self-determined Research Funds of CCNU from the Colleges’ Basic Research and Operation of MOE (No. CCNU24JC033).} 
	
\thanks{Zhifeng~Wang and Longlong Li are with the Faculty of Artificial Intelligence in Education, Central China Normal University, Wuhan 430079, China.}

\thanks{Minghui Wang is with the CCNU Wollongong Joint Institute, Central China Normal University, Wuhan 430079, China.}

\thanks{Chunyan Zeng is with the Hubei Key Laboratory for High-Efficiency Utilization of Solar Energy and Operation Control of Energy Storage System, Hubei University of Technology, Wuhan 430068, China (Corresponding authors: Zhifeng Wang, Chunyan Zeng, E-mail: zfwang@ccnu.edu.cn, cyzeng@hbut.edu.cn).}

}

\markboth{arxiv, 2024}%
{Shell \MakeLowercase{\textit{et al.}}: A Sample Article Using IEEEtran.cls for IEEE Journals}


\maketitle

\begin{abstract}
The integration of Artificial Intelligence (AI) into the modern educational system is rapidly evolving, particularly in monitoring student behavior in classrooms—a task traditionally dependent on manual observation. This conventional method is notably inefficient, prompting a shift toward more advanced solutions like computer vision. However, existing target detection models face significant challenges such as occlusion, blurring, and scale disparity, which are exacerbated by the dynamic and complex nature of classroom settings. Furthermore, these models must adeptly handle multiple target detection.
To overcome these obstacles, we introduce the Student Learning Behavior Detection with Multi-Scale Deformable Transformers (SCB-DETR), an innovative approach that utilizes large convolutional kernels for upstream feature extraction, and multi-scale feature fusion. This technique significantly improves the detection capabilities for multi-scale and occluded targets, offering a robust solution for analyzing student behavior. SCB-DETR establishes an end-to-end framework that simplifies the detection process and consistently outperforms other deep learning methods. Employing our custom Student Classroom Behavior (SCBehavior) Dataset, SCB-DETR achieves a mean Average Precision (mAP) of 0.626, which is a 1.5\% improvement over the baseline model's mAP and a 6\% increase in AP50. These results demonstrate SCB-DETR's superior performance in handling the uneven distribution of student behaviors and ensuring precise detection in dynamic classroom environments. The source code of this study is publicly available at \href{https://github.com/CCNUZFW/SCB-DETR}{https://github.com/CCNUZFW/SCB-DETR}.
\end{abstract}


\section{Introduction}


\IEEEPARstart{T}{he} rapid advancement of AI stands as a cornerstone of contemporary technological innovation \cite{Liao2024,Ma2023b,Wang2022as,Min2019}, with computer vision being a particularly transformative area within AI \cite{Zhao2024b,Wang2015a,Zeng2023b,Li2023h,Wang2021,Tian2018,Min2018,Wang2017}. This field enables machines to interpret visual data, fundamentally altering their interaction with the environment and fostering novel applications across various sectors \cite{Goldblum2023}. The domain of education has not been exempt from this technological wave \cite{Wang2024b,Dong2024,Wang2023j,Li2023i,Wang2023d,Li2023g,Lyu2022,Li2023f}. Here, AI is increasingly integral, evidenced by the proliferation of smart education solutions and the concept of smart classrooms \cite{Judijanto2024,Wang2025,Wang2024m}. These advancements are not merely enhancing traditional educational methodologies but are also setting the stage for future educational frameworks that are more adaptive, interactive, and efficient \cite{Wang2023g,Wang2023w,Wang2023l,Wang2022at}.

The application of these emerging educational technologies is anticipated to address numerous shortcomings inherent to the traditional education model \cite{Dong2024}. One such issue is the rigid, one-size-fits-all teaching approach that often overlooks individual student needs and potential \cite{Lee2024}. Furthermore, traditional methods struggle to consistently engage and motivate students \cite{Wang2023d}. The integration of AI is progressively mitigating these challenges by introducing more adaptive and personalized educational experiences. In terms of teaching methods, machine learning-based recommender systems are pivotal in tailoring the educational journey. These systems enhance the learning experience by providing personalized learning paths, thereby promoting efficient knowledge acquisition \cite{bib35}. In the realm of content delivery, data mining technologies are instrumental \cite{Wang2023j}. They aid educators in gaining insights into students' learning habits, mastery of content, and the effectiveness of instructional materials through the analysis of extensive educational data \cite{bib36}. However, the assessment of student progress remains a challenge. Traditional methods, which often rely solely on teacher observation, do not provide a comprehensive view of student capabilities \cite{Zhao2023a}. This gap highlights the necessity for sophisticated target detection models tailored to the educational context \cite{Abichandani2023}, which is characterized by its complexity and variability \cite{Wu2023g,Poonja2023}. General models are inadequate for addressing the specific needs of such environments, leading to the exploration of specialized models that can handle issues like occlusion and varying target scales \cite{Wang2023g}. Inspired by the advanced deep learning technology \cite{Zeng2024g,Wang2023f,Zeng2024f,Chen2023b,Zeng2024c,Wang2022t,Zeng2022a,Wang2021m,Zeng2024d,Wang2020h,Zeng2024,Wang2018a,Zeng2024a,Zhang2024i,Zeng2023a,Zeng2023,Zeng2021a,Zeng2021b,Zeng2020,Zeng2018}, our research thus centers on developing student learning behavior detection models specifically designed for classroom settings, addressing these unique challenges.


Additionally, our research identified that the task of detecting students' classroom behaviors is complicated by the diverse and interactive nature of these behaviors. Student interactions often create dynamic and complex behavioral patterns that are challenging to capture and analyze with standard models. To address these complexities effectively, we introduce a novel neural network framework in this paper, named SCB-DETR. This framework synergizes the strengths of traditional Convolutional Neural Networks (CNNs) \cite{Gu2018,Zeng2024e,Zeng2023c,Wang2023a,Zeng2022,Zeng2021c,Zeng2020a} and Transformers \cite{bib2,Zheng2024,Zeng2024b}. It features an extended convolutional kernel size and incorporates a multi-scale fusion architecture, enhancing its capability for targeted detection and classification of student behaviors.
We have developed a novel Large-scale Convolutional Kernel Backbone Network, termed LKNeXt, designed specifically for efficient upstream characterization of student behaviors. Furthermore, we introduce the Hybrid Feature Fusion Architecture (HFFA), which significantly enhances the model's detection capabilities. The contributions of our work are detailed as follows:

\begin{enumerate}
	\item \textbf{Enhanced Feature Extraction with LKNeXt Backbone}: We have developed a novel backbone network, LKNeXt, which utilizes an enlarged convolutional kernel. This design significantly enhances the model's ability to extract upstream features related to student behavior, improving both the accuracy and efficiency of behavioral analysis in classroom settings.
	
	\item \textbf{Development of HFFA for Multi-Scale Feature Fusion}: Our work introduces the HFFA, a crucial innovation within our model. HFFA integrates the strengths of feature pyramids and encoder-based feature extraction to enrich the model’s understanding of spatial relationships and focus perception within the learning environment. This architecture facilitates a deeper and more nuanced representation of behavior features, crucial for detecting and analyzing occlusion student behavior.
	
	\item \textbf{Creation and Release of SCBehavior Dataset}: We have compiled and made publicly available the SCBehavior dataset, which is specifically tailored for the task of recognizing common student behaviors in classroom settings. This dataset encompasses seven typical behaviors, providing a robust foundation for training and validating our proposed model. The SCBehavior dataset can be accessed at \href{https://github.com/CCNUZFW/SCBehavior}{https://github.com/CCNUZFW/SC\-Behavior}.
\end{enumerate}


The structure of this paper is organized to clearly present our research and findings. Section \ref{RW} reviews the related work, providing context and background to the advancements in AI-driven educational tools and specific challenges in behavior detection. Section \ref{MED} details our proposed methodology, introducing the SCB-DETR framework along with the innovative LKNeXt and HFFA architectures. Section \ref{EXP} presents the experimental setup, discusses the results obtained, and engages with the implications of these findings. Finally, Section \ref{CON} concludes the paper by summarizing the key outcomes and suggesting avenues for future research.

\section{Related Work} \label{RW}

This section offers a comprehensive review of the methodologies currently employed for identifying student learning behaviors within classroom settings, alongside recent advancements in target detection technologies and detection Transformer. It also provides an analysis of the backbone network structures that are frequently utilized in this research area.

\subsection{Student Learning Behavior Analysis}

The concept of the smart classroom has evolved rapidly over recent years, leading to significant advancements in deep learning-based methods for detecting student behaviors in classroom settings. This subsection explores several key studies that have contributed to this field.

Wang et al. \cite{bib3} utilized the R-FCN architecture to detect yawning behaviors among students, incorporating pruning operations and a novel mouth fitting method to enhance detection rates and achieve precise behavior analysis in classrooms. Zhang et al. \cite{bib4} introduced a model that employs coordinate-based attention combined with fused residual networks. This model significantly improves the accuracy of classroom behavior detection by effectively utilizing channel and spatial information to highlight pertinent features in student images.

Further contributions include those by Wang et al. \cite{bib5}, who designed a novel network using YOLO as the baseline. By creating their own student behavior dataset and applying data enhancement techniques, they achieved improved detection outcomes. Additionally, Wang et al. \cite{bib6} advanced the field by integrating a channel attention mechanism (SE) into their model, thereby enhancing the perception of student behaviors in classroom environments.

\textbf{Summary}: These studies illustrate the ongoing evolution of techniques for detecting student behaviors, yet they also highlight a gap in the application of end-to-end models within educational settings. Addressing this, our work introduces a novel end-to-end target detection model specifically tailored for the complexities of classroom student behavior analysis.

\subsection{Object Detection}

Object detection algorithms, a cornerstone of computer vision, are commonly divided into two main categories: two-stage and one-stage detectors. The two-stage detectors, such as R-CNN \cite{bib26}, marked the initial application of deep learning in object detection. R-CNN uses selective search to initially extract candidate regions from an image, which are then classified using a convolutional neural network. Building upon this, Fast R-CNN \cite{bib27} and Faster R-CNN\cite{bib14} were introduced to enhance the speed and efficiency of the original R-CNN. Notably, Faster R-CNN, appearing in 2015, incorporated Region Proposal Networks that significantly accelerated the detection process by enabling the simultaneous prediction of object bounds and scores for each object category.

In contrast, one-stage detectors such as YOLO simplified the detection process by eliminating the proposal stage, directly predicting object classifications and localizations from full images. This approach offers faster processing times but traditionally suffered from lower accuracy compared to two-stage methods. Enhancements to this approach were realized through developments in algorithms like SSD \cite{bib16} and RetinaNet \cite{bib13}, which incrementally improved both speed and accuracy, influencing subsequent iterations of YOLO.

\textbf{Summary}: The advent of the Transformer architecture introduced a novel paradigm in object detection, leveraging self-attention mechanisms to enhance the detection capabilities further. This evolution underscores the continuous innovation in the field, seeking to balance the trade-offs between detection speed and accuracy.

\subsection{Detection Transformer}

Detection Transformer (DETR) \cite{bib32} represents a transformative approach in object detection by treating it as an ensemble prediction problem. This end-to-end model simplifies the detection pipeline by eliminating the need for traditionally hand-designed components such as non-maximal suppression (NMS) and anchor generation algorithms, which typically encode prior knowledge about the detection task \cite{bib7}. The DETR model comprises three main components: the backbone network, the encoder-decoder structure, and the prediction head. While the attention mechanism within the encoder-decoder framework enhances target detection, it introduces challenges such as high computational demands, extensive training times, and suboptimal performance in detecting small targets.

To address these issues, subsequent innovations have emerged. Deformable-DETR \cite{bib8} leverages the sparse spatial sampling capability of deformable convolution to reduce computational complexity and accelerate model convergence. This adaptation also improves the detection performance for smaller targets, addressing one of the key limitations of the original DETR.
DINO \cite{bib10} enhances the DETR model by integrating a denoising training approach and a hybrid query selection method for anchor initialization. Additionally, a double forward pass technique for frame prediction significantly quickens the convergence rate. These improvements not only boost performance but also reduce both the model size and the amount of required pre-training data.

\textbf{Summary}: Despite these advancements, a specialized end-to-end model for detecting student behaviors in classroom settings has yet to be developed. In response, this paper introduces a novel framework based on Deformable-DETR, specifically tailored for classroom student behavior detection.

\subsection{Backbone Network}

In deep learning, the backbone network is crucial, serving as the foundational component for constructing neural network models. It is primarily responsible for extracting feature representations from input data, playing a pivotal role in tasks such as image recognition, object detection, and semantic segmentation. In computer vision tasks, the backbone network processes raw images and extracts essential features, which are then passed to task-specific modules for further processing.

The evolution of backbone networks has seen several significant developments. Developed by \cite{LeCun1998} for recognizing handwritten digits, LeNet was among the first to utilize alternating convolutional and pooling layers for feature extraction, followed by classification through fully connected layers. While pioneering, its limitations in depth and design restricted its application to more complex tasks.
In the 2012 ImageNet competition, \cite{Krizhevsky2012} introduced AlexNet, which expanded upon LeNet by incorporating Rectified Linear Units (ReLU) to combat vanishing gradients during training. This model required substantial computational resources and implemented strategies to prevent overfitting, setting a new standard in CNN architectures.
Developed by \cite{Simonyan2015}, VGG improved upon AlexNet by reducing the size of convolutional kernels and increasing network depth, which enhanced accuracy and generalization capabilities.
In 2015, \cite{He2016} introduced ResNet, which addressed deep network training challenges by incorporating residual learning. This allowed for the training of much deeper networks by facilitating better gradient flow.
Introduced by \cite{Tan2019}, EfficientNet optimized the balance between network depth, width, and resolution, enhancing performance and computational efficiency.
The Efficient Long-range Attention Network (ELAN), introduced by \cite{Zhang2022ao}, features the Group-wise Multi-scale Self-Attention (GMSA) module, which manages self-attention across various feature scales, enhancing the handling of long-range dependencies with reduced computational demands. Current research on the backbone is facing significant challenges in complex environments, such as educational settings \cite{bib22,bib23}.

\textbf{Summary:} Despite these advancements, there is a notable gap in the adaptation of backbone networks for educational applications. Our research aims to fill this void by developing a backbone network specifically tailored for educational environments.

\section{Proposed Method} \label{MED}

\begin{figure*}[htbp]
	\centering
	\includegraphics[width=1\textwidth]{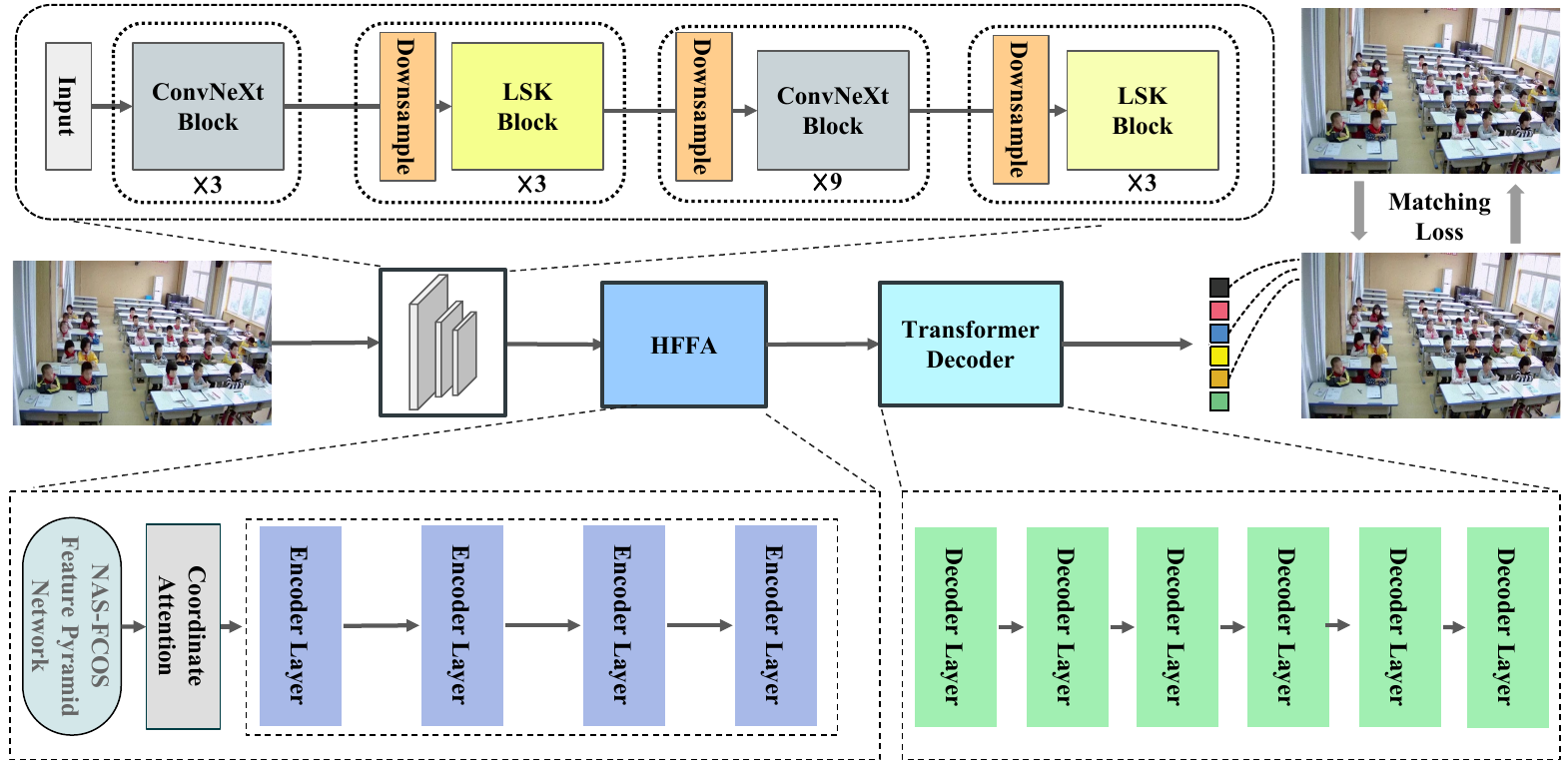}
	\caption{SCB-DETR general framework: student classroom related images are first input into the LKNeXt backbone constructed from a combination of two types of large convolutional blocks for upstream feature extraction, thus outputting four-dimensional features. Subsequently, the four-dimensional features are fed into the HFFA framework for feature cross-fertilisation, and finally the decoder uses the feature representation of HFFA to generate predictions for target detection.}
	\label{fig1}
\end{figure*}

The SCB-DETR framework is comprised of several key components: a feature extraction network, the HFFA architecture, a decoder, and a prediction head. The workflow begins with the LKNeXt backbone, which is integral to the feature extraction network. This backbone processes the input images and extracts complex features, which are subsequently structured into four-dimensional feature maps.

These feature maps are then passed into the NASFCOS-FPN component of the HFFA architecture. Following this, a normalization-based coordinate attention mechanism is applied to the features. This attention mechanism is particularly designed to enhance the encoding of location information within the feature maps, crucial for accurate target detection.

The enriched features proceed to the next phase of the HFFA framework, where four encoders equipped with a multi-scale deformable attention mechanism are utilized. This configuration is specifically designed to sparsify the features and enhance the detection capability across multiple scales. This aspect of the architecture is essential for handling the complexities associated with varying object sizes within the images.

Finally, the refined features are relayed to the decoder and the prediction head. The decoder, similar to that used in DETR, processes these features and employs a Feed-Forward Network (FFN) to produce the final classification outcomes and object bounding boxes. This process is visualized in Fig. \ref{fig1}, which illustrates the comprehensive structure of our network model.

\subsection{LKNeXt Backbone}

In the context of student classroom behavior, which presents unique environmental and behavioral characteristics, the necessity for effective feature extraction and computational efficiency is paramount. To address this, we developed the LKNeXt, a large convolutional kernel-based backbone network. This network is specifically designed to optimize the capture of nuanced student behaviors in classroom settings and to distinguish similar behaviors influenced by nearby students.

The LKNeXt backbone, embedded within our overall model's feature extraction network, incorporates large convolutional blocks that are enhanced with residual connections. This design decision is inspired by ResNet, which effectively mitigates the problem of gradient vanishing in deep networks through residual connectivity, thereby enriching feature extraction by maintaining robust information flow across the network layers. Our adaptation employs these principles to ensure that even with deep network structures, the LKNeXt can extract detailed and significant features without a loss in performance.

The motivation behind the development of LKNeXt stems from the challenges posed by the diversity of classroom environments, particularly when dealing with uniformly dressed students in primary and secondary schools. Differentiating and extracting distinct behavioral features in such scenarios is crucial. LKNeXt excels in this regard, offering superior performance compared to traditional CNNs at the same parameter scale. Furthermore, it significantly expands the sensory field of the model, enhancing the accuracy of behavior recognition and environmental perception.

The core innovation of LKNeXt lies in its utilization of large convolutional kernels. These kernels are adept at integrating extensive spatial information in a single convolutional operation, making them particularly effective in capturing subtle spatial dynamics in student behaviors, such as hand raising, reading, or interacting with peers. This ability contrasts sharply with traditional smaller kernels, which may fail to capture such nuances effectively.

To construct LKNeXt, we drew inspiration from seminal works on large convolutional kernels, leading to a backbone network structure that consists of two distinct types of blocks, each optimized for specific feature integration and processing tasks within the educational environment.

\subsubsection{ConvNeXt Block}

\begin{figure}[h]
	\centering
	\includegraphics[width=0.45\textwidth]{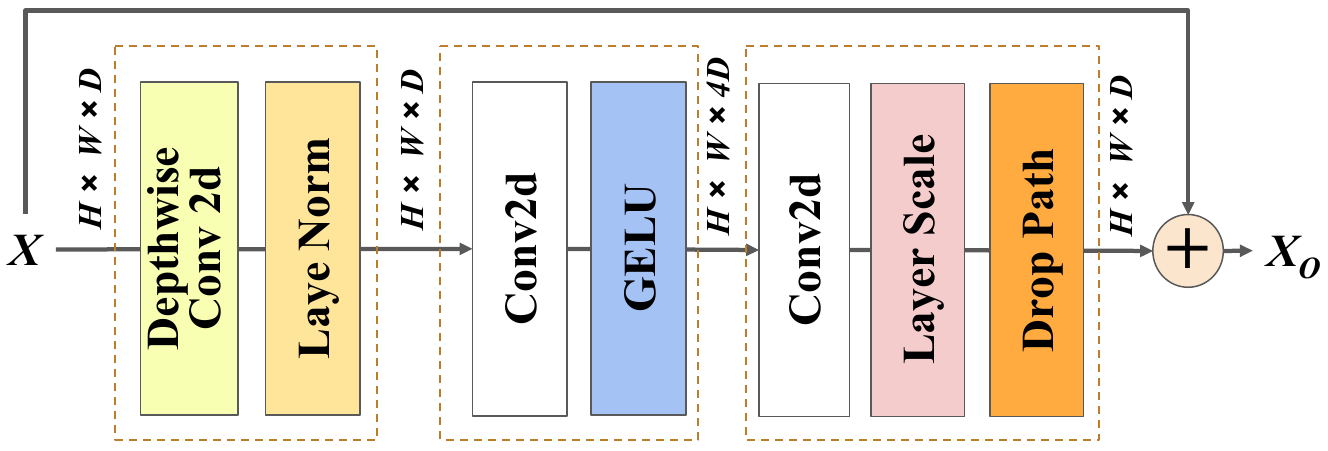}
	\caption{The structure of the ConvNeXt block.}
	\label{fig2}
\end{figure}

The ConvNeXt Block is a pivotal component of our LKNeXt backbone, designed specifically to handle the unique challenges presented by student classroom behavior analysis. The processing begins with a depth-separable convolution applied to the input features of student classroom behaviors. This method significantly reduces the number of covariates, thereby enhancing computational efficiency. Following this, a Layer Normalization (LayerNorm) is applied to the feature maps to improve training stability and efficiency while simultaneously minimizing internal covariate bias, as shown in Fig. \ref{fig2}.

The architecture then incorporates two \(1 \times 1\) convolutions paired with a GELU activation function \cite{bib24}. This configuration constructs an Inverted Bottleneck structure, which is effective in reducing the computational load while maintaining essential feature information. The sequence is completed with residual concatenation and a Drop path strategy, which are instrumental in enhancing the generalization capabilities of the network and combating the gradient vanishing problem often encountered in deep neural networks.

The operations within the ConvNeXt Block can be formally expressed with the following equations:

\begin{equation}
	X' = PWConv(GELU(PWConv(LN(DWConv(X)))))
\end{equation}

\begin{equation}
	X_O = X + X'
\end{equation}

This formulation ensures that the output not only retains but also enhances the original input features through the block's operations, contributing to more robust and accurate behavior detection.

\subsubsection{LSK Block}

The LSK Block in our LKNeXt backbone is designed to meticulously handle the extraction of nuanced features from student behaviors in classroom settings, utilizing a dual-convolution approach. Initially, we deploy two distinct large convolution operations: a \(5 \times 5\) grouped convolution and a \(7 \times 7\) spatial expansion convolution. These convolutions are aimed at extracting features from diverse spatial dimensions to broaden the sensory field and enhance the feature detail of individual channels, surpassing the capabilities of standard convolutions.

After extracting these features, they are combined and refined as follows:

\begin{equation}
	\widehat{SA} = Conv([SA_{avg};SA_{max}])
\end{equation}

This step involves merging the downscaled features using a \(1 \times 1\) convolution to perform feature dimensionality reduction, followed by computing the mean and maximum values, which help in emphasizing both the common and unique aspects of the extracted features.

Subsequently, the fused features are refined using an attention mechanism, initially applying a sigmoid function to generate attention weights:

\begin{equation}
	\widetilde{SA} = \sigma(\widehat{SA})
\end{equation}

The attention weights are then used to adjust the output from the two feature branches, enhancing feature selectivity and relevance. The number of channels is restored through a \(1 \times 1\) convolution layer, and the final output is derived by applying a weighted sum of these features to the input features, maintaining a residual connection. This ensures the integration of new and original features, preserving essential information while adding depth to the feature analysis.

\begin{figure}[h]
	\centering
	\includegraphics[width=0.49\textwidth]{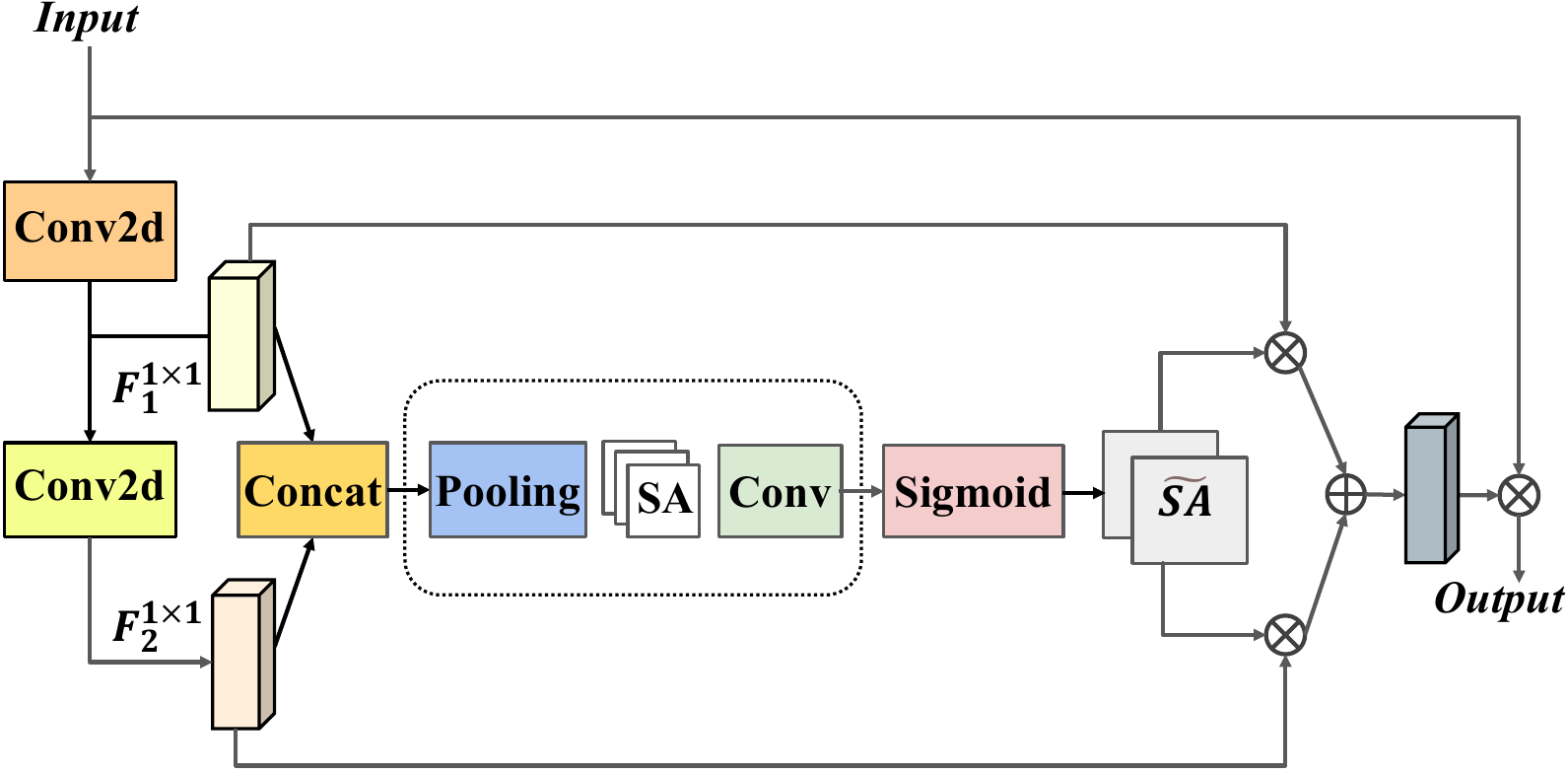}
	\caption{The structure of the LSK block.}
	\label{fig3}
\end{figure}

The output then progresses to further large convolution layers for additional feature enhancement. The detailed structure of the LSK Block is illustrated in Fig. \ref{fig3}, demonstrating its integral role in feature extraction and refinement within the network.

\subsection{Hybrid Feature Fusion Architecture (HFFA)}

The HFFA is specifically tailored to optimize the feature extraction and fusion processes for student behavior detection. This architecture integrates a multi-scale feature pyramid with a deformable encoder, diverging from the conventional six-layer encoder structure typically used in such applications. The design of HFFA is particularly effective in aggregating behavioral features at various scales and extracting fine-grained characteristics through advanced cross-fusion techniques.

The incorporation of an attention mechanism is a central aspect of the HFFA. It has become a focal point in current research due to its ability to enhance network performance significantly. In the context of student behavior detection, attention manipulation is often employed to isolate and amplify distinctive features of classroom student behavior. This approach is instrumental in addressing common challenges in the field, such as variability in classroom settings, differences in student sizes, and behavioral occlusions.

The initial motivation for developing the HFFA stemmed from the specific needs of student behavior analysis in classroom environments, where behaviors may not only be similar among students but also influenced by the actions of nearby peers. To adequately capture these complex behavioral interactions, HFFA incorporates an attention module that enables the acquisition of global information and facilitates effective feature extraction. This integration ensures that the architecture can handle the nuanced dynamics present in classroom settings, thereby improving the accuracy and reliability of behavior detection.

\subsubsection{NASFCOS-FPN}

\begin{figure}[h]
	\centering
	\includegraphics[width=0.49\textwidth]{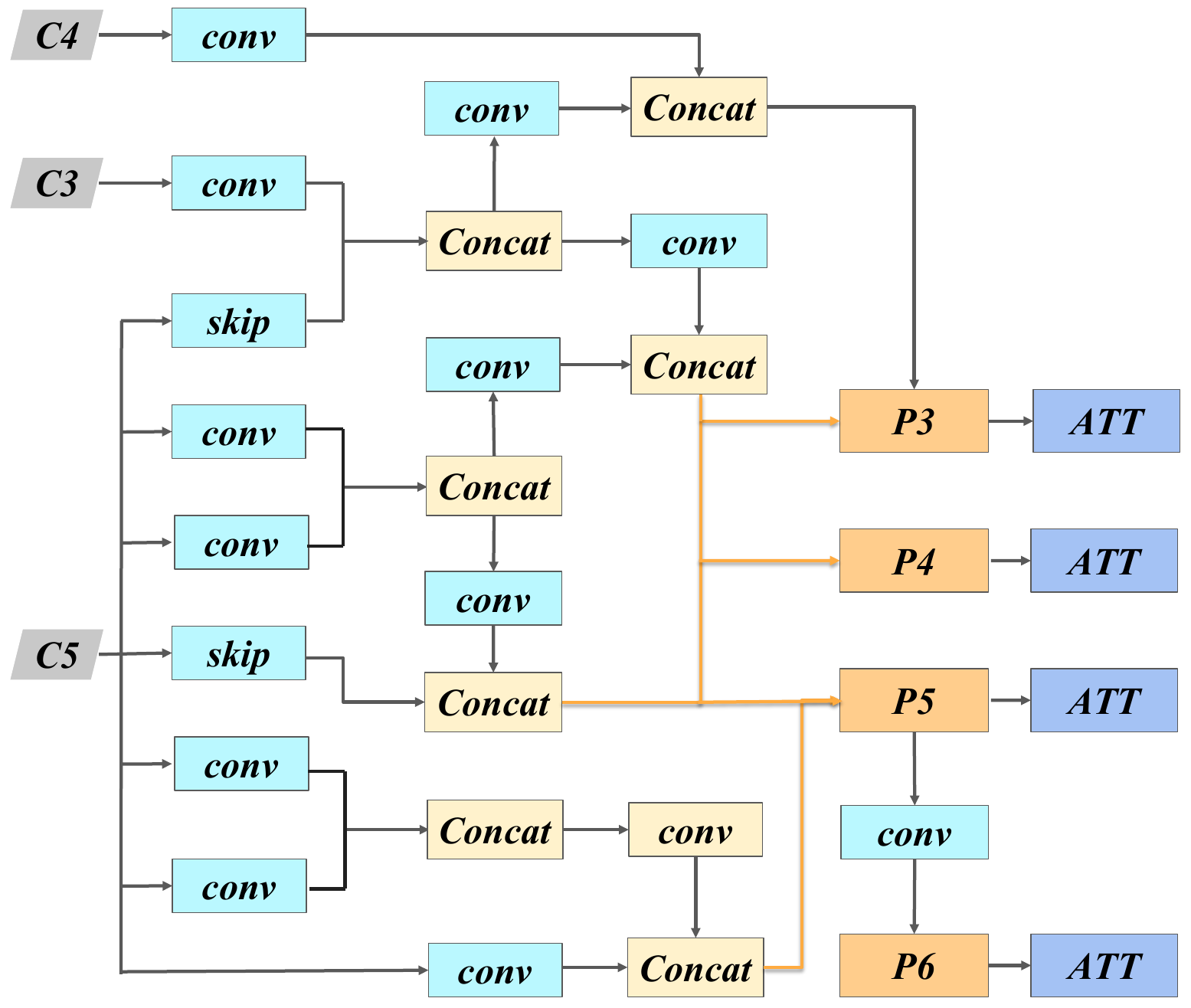}
	\caption{The framework of attention-based NASFCOS-FPN.}
	\label{fig4}
\end{figure}

Feature Pyramid Networks (FPN) represent a significant advancement in neural network architecture designed to enhance object detection performance in computer vision tasks. The core principle of FPN is to leverage the inherent multi-scale, pyramidal hierarchy of feature layers in Convolutional Neural Networks and to fuse these layers to improve the detection accuracy of objects across different sizes. The NASFCOS-FPN extends this concept by incorporating a coordinate attention mechanism, which is detailed in Fig. \ref{fig4}.

In the NASFCOS-FPN structure, initial features from the backbone network are first enriched and diversified through a sophisticated fusion process within the FPN. Subsequently, the Coordinate Attention mechanism is applied. This mechanism is specifically designed to enhance the model's sensitivity to spatial distributions and critical areas of student behaviors within the classroom. It effectively emphasizes key features that are crucial for accurate behavior recognition.

For the feature pyramid module of NASFCOS-FPN, we begin by standardizing the input features to a uniform channel count using an adaptive convolution (Adaptconv). This standardization is crucial for maintaining consistency in feature processing. Following this, a series of \(1 \times 1\) and \(3 \times 3\) convolutions, which share the same architectural structure, are employed to execute feature fusion and downsampling. This methodology not only maintains the integrity of the feature information but also enhances it, ensuring that the pyramid captures a comprehensive representation of the input features across different scales.

\subsubsection{Coordinate Attention Mechanism}

Recent advancements in attention mechanisms, particularly channel attention, have demonstrated significant improvements in model performance across various applications. However, these methods often overlook the critical aspect of location information, which is essential in applications like classroom student behavior detection. To address this, we have integrated a Coordinate Attention (CA) module following the feature pyramid in our model. This integration aims to enable the attention module to capture remote spatial interactions with precise location information by encoding the horizontal and vertical coordinates of each channel individually.

\begin{figure}[h]
	\centering
	\includegraphics[width=0.49\textwidth]{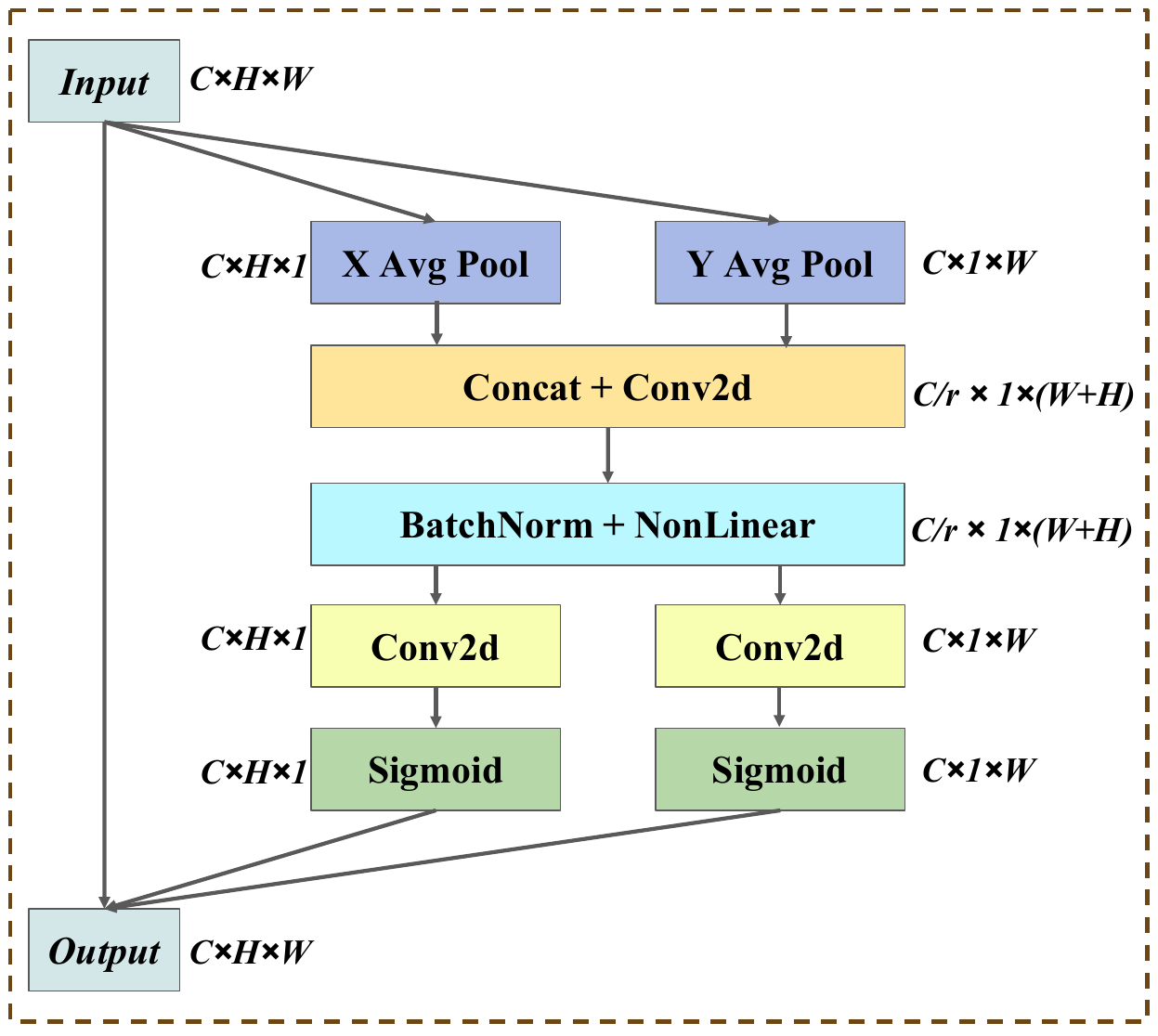}
	\caption{The structure of Coordinate Attention.}
	\label{fig5}
\end{figure}

The typical approach of using global pooling for spatial information encoding in channel attention mechanisms compresses the entire spatial dimension into channel descriptors \cite{bib18}. While effective in reducing dimensionality and computational load, this approach significantly limits the model's ability to maintain accurate location information, which is paramount for detecting nuanced student behaviors in varied classroom settings.

To overcome these limitations and enhance the model's spatial sensitivity, the CA module decomposes the global pooling process into separate one-dimensional encoding operations for each spatial dimension. This method preserves more granular location data and allows the model to better understand and react to the spatial dynamics of classroom environments. By maintaining precise location information, the Coordinate Attention mechanism can effectively highlight areas of interest and improve the overall detection accuracy of student behaviors, as shown in Fig. \ref{fig5}.

For an input \( X \), the Coordinate Attention mechanism refines the traditional global pooling process by splitting it into two distinct one-dimensional feature encoding operations. This adjustment allows for more detailed spatial analysis and better retention of location-specific information, crucial for precise behavior detection in complex environments like classrooms. The operations are defined as follows:

\begin{equation}
	z_{c}^h(h) = \frac{1}{W}\sum_{0 \leq i < W} x_c(h,i).
\end{equation}

\begin{equation}
	z_{c}^w(w) = \frac{1}{H}\sum_{0 \leq j < H} x_c(j,w).
\end{equation}

The first equation computes a horizontal pooling by aggregating features across the width \( W \) of the feature map for each vertical position \( h \), producing a feature vector that captures horizontal spatial dependencies. Conversely, the second equation performs a vertical pooling across the height \( H \) for each horizontal position \( w \), creating a complementary feature vector that encapsulates vertical spatial dependencies.

These two transformations yield a pair of direction-aware feature maps, which are fundamentally different from the outputs of typical SE (Squeeze-and-Excitation) attention mechanisms that generate a single, globally aggregated feature vector. By maintaining separate feature vectors for each dimension, the Coordinate Attention mechanism can effectively capture long-range dependencies along one dimension while preserving critical location information along the other. This dual approach enhances the model’s ability to precisely localize and respond to diverse and nuanced student behaviors within the spatially complex context of a classroom.
 
Once the one-dimensional feature encodings are computed, the next step in the Coordinate Attention mechanism involves concatenating the transformed features \( z^h \) and \( z^w \). This concatenated vector is then subjected to further transformation using a convolutional function, where \( F_1 \) represents a \(1 \times 1\) convolution, and \( \delta \) denotes a nonlinear activation function. This transformation is captured by the following equation:

\begin{equation}
	f = \delta(F_1([z^h, z^w])).
\end{equation}

The output \( f \) serves as an intermediate feature map that combines both horizontal and vertical spatial information, which is crucial for maintaining spatial continuity and detail. Subsequently, this feature map \( f \) is split into two separate feature vectors, which are further refined using two additional \(1 \times 1\) convolutions followed by a sigmoid activation function to normalize the output. These operations are expressed in the equations below:

\begin{equation}
	g^h = \sigma(F_h(f^h)),
\end{equation}

\begin{equation}
	g^w = \sigma(F_w(f^w)).
\end{equation}

These two feature vectors \( g^h \) and \( g^w \) represent attention maps that modulate the input features by applying distinct weights to different spatial dimensions, thereby enhancing the model’s ability to focus on relevant spatial regions. The final output of the Coordinate Attention mechanism is computed by applying these attention weights back to the original input features, effectively allowing the model to emphasize or de-emphasize specific areas based on the learned importance. The combination of these weights with the original input features \( x_c \) is described by the final equation:

\begin{equation}
	y_c(i,j) = x_c(i,j) \times g_c^h(i) \times g_c^w(j).
\end{equation}

This final operation produces an output \( y_c \) that integrates attention-modulated features across both spatial dimensions, ensuring that the model's predictions are sensitive to the contextual nuances of the input data.

\subsubsection{Encoder-Decoder with Deformable Attention Mechanisms}

\begin{figure}[h]
	\centering
	\includegraphics[width=0.45\textwidth]{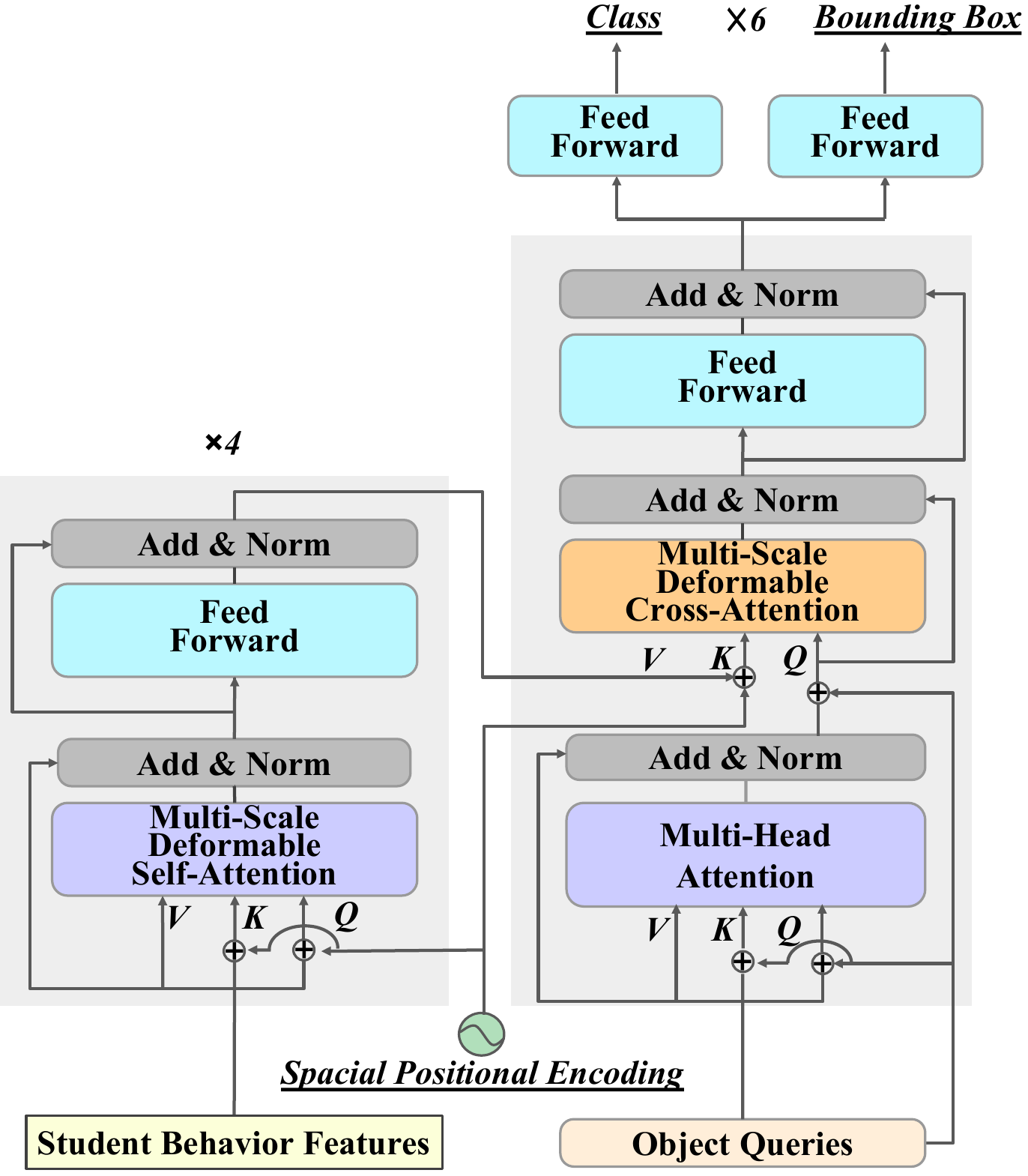}
	\caption{Encoder-Decoder with Deformable Attention.}
	\label{fig6}
\end{figure}

In our architecture, the features that have been enriched and fused within the feature pyramid are flattened into a one-dimensional feature sequence. This sequence is combined with positional encodings to preserve the spatial relationships before being fed into the Encoder for global feature modeling. The integration of an improved deformable attention mechanism, as utilized in Deformable-DETR, is crucial for learning the relationships between different features by allowing for adaptive feature sampling and focusing on relevant parts of the input data, as shown in Fig. \ref{fig6}.

The backbone of our encoder-decoder architecture is inspired by the Transformer model, originally developed for machine translation tasks. The core principle of the Transformer architecture involves the Multi-Head Attention mechanism. Given a query element and a set of key elements, this mechanism computes attention weights that measure the compatibility of the query with each key. These weights are then used to adaptively aggregate content from the keys, enabling the model to focus on relevant information dynamically. 

To enhance the model's ability to attend to different features from various representation subspaces and spatial locations, the outputs of multiple attention heads are combined using linear aggregation with learnable weights. This allows each head to specialize in different aspects of the feature space, which increases the flexibility and power of the model to capture complex dependencies within the data. The deformable component of the attention mechanism further refines this process by focusing on a subset of positions according to the data's structural characteristics, thereby enhancing efficiency and model performance in scenarios with high variability, such as classroom behavior detection.

In our model, we define a query element indexed by \( q \) from the set \( \Omega_q \) with a feature representation \( z_q \in \mathbb{R}^C \), and a key element indexed by \( k \) from the set \( \Omega_k \) with a feature representation \( x_k \in \mathbb{R}^C \), where \( C \) denotes the dimension of the feature space. The sets \( \Omega_q \) and \( \Omega_k \) specify the collections of query and key elements, respectively. The multi-head attention mechanism is then computed as follows:

\begin{equation}
	\text{MultiHeadAttn}(z_q, x_k) = \sum^M W_m \left[\sum_{w0} A_{mqk} \times W_m' x_k\right]
\end{equation}

Here, \( M \) represents the number of attention heads. The matrices \( W_m' \in \mathbb{R}^{C_v \times C} \) and \( W_m \in \mathbb{R}^{C \times C_v} \) are sets of learnable weights specific to each attention head. The attention weights \( A_{mqk} \) are calculated and normalized according to the softmax function over the scaled dot-product of the query and key features, facilitated by additional learnable transformations \( U_m \) and \( V_m \):

\begin{equation}
	A_{mqk} \propto \exp\left(\frac{Z_q^T U_m^T V_m x_k}{\sqrt{C_v}}\right)
\end{equation}

This formulation ensures that the attention mechanism adaptively focuses on the most relevant features by weighting the input key features \( x_k \) according to their relevance to the query feature \( z_q \), enhanced by the multi-head structure that allows the model to attend to information from different representation subspaces at different positions.

Transformers typically apply attention uniformly across all spatial locations in an image feature map. However, this approach can be computationally inefficient and less effective when specific spatial focus is required. To address this, Deformable-DETR incorporates a deformable attention module inspired by deformable convolution techniques \cite{bib12}. This module selectively focuses on a limited set of key sampling points around a predetermined reference point, thus ignoring irrelevant spatial areas. Such a design enhances model convergence and improves feature spatial resolution by concentrating computational resources on the most informative parts of the feature map.

Given an input feature map \( x \), and defining \( q \) as the index for a query element with content feature \( z_q \) and a 2D reference point \( p_q \), the deformable attention feature is computed as follows:

\begin{equation}
	\text{DeformAttn}(z_q, p_q, x) = \sum^{M}_{m=1}W_m\left[\sum_{k=1}^{K}A_{mqk} \cdot W_m'x(\Delta p_{mqk})\right]
\end{equation}

In this formula, \( m \) indexes the attention heads, and \( k \) indexes the sampled keys, where \( K \) is the total number of keys. The term \( \Delta p_{mqk} \) denotes the sampling offset for the \( k^{th} \) key in the \( m^{th} \) attention head, and \( A_{mqk} \) is the corresponding attention weight.

This mechanism allows the deformable attention module to dynamically adjust the focus of the model to regions of the image that are most relevant to the task at hand. By limiting the number of spatial locations each query attends to, the deformable attention efficiently processes large images and complex scenes, significantly reducing the computational burden while maintaining or even enhancing the model's performance.

The deformable attention module is specifically engineered to optimize convolutional feature mapping, drawing on principles from contemporary target detection frameworks. Its design is particularly effective in addressing the unique challenges posed by different scales within an image, making it a versatile component in our detection system.

To enhance its applicability across varying scales of input data, the deformable attention has been adapted for multiscale applications. This multiscale deformable attention mechanism operates similarly to its single-scale counterpart but is designed to handle a more complex feature landscape. Instead of sampling \( K \) points from a single-scale feature map, it samples \( LK \) points from a multiscale feature map. This adjustment allows the module to capture more comprehensive spatial information, accommodating variations in object size and contextual detail across the feature map.

This capability to dynamically adjust to different scales and focus on the most informative parts of the image at each scale makes the deformable attention module particularly effective in scenarios where the relevance and scale of features can vary significantly, such as in environments with diverse object sizes or varying distances from the camera.

\subsection{Optimization and Training}

The performance of our model hinges significantly on the choice and implementation of the loss function, which quantifies the discrepancy between the predicted outputs and the actual labels or values. Our model employs a combination of classification loss, bounding box regression loss, and Intersection over Union (IoU) loss to optimize its performance across various aspects of target detection.

\begin{enumerate}
	\item \textbf{Classification Loss:} This component of the loss function assesses the accuracy of the model in predicting the correct category of the targets. It calculates the difference between the predicted probability distribution across various categories and the actual category labels, guiding the model to improve its predictive accuracy.
	
	\item \textbf{Bounding Box Regression Loss:} Critical for the precise localization of objects, this loss measures the discrepancy between the coordinates of the predicted bounding box and the actual bounding box. The bounding boxes are typically defined by the coordinates of their center points, width, and height. Minimizing this loss is crucial for enhancing the model’s ability to accurately locate objects.
	
	\item \textbf{Intersection over Union (IoU) Loss:} Also known as Jaccard's index, IoU loss evaluates the extent of overlap between the predicted bounding box and the true bounding box. It is a crucial metric for determining how well the model's predictions align with the actual object locations.
\end{enumerate}

Our objective is to simultaneously minimize these losses to boost the model's capability in both recognizing the categories of objects and pinpointing their positions. The combined approach to these distinct but complementary loss functions ensures a holistic enhancement of the model's performance in the complex task of target detection.

For the task of classroom student behavior recognition, the diversity in student behaviors and physical sizes across various classroom locations poses unique challenges. To effectively address these challenges, particularly for multi-scale object detection, our model employs a sophisticated combination of loss functions tailored to enhance performance across different dimensions of the task.

We utilize Focal Loss for classification to dynamically adjust the importance of correctly classifying each sample, prioritizing harder-to-classify instances which typically have a greater impact on the model's performance. For bounding box regression, we combine Smooth L1 Loss and Generalized Intersection over Union (GIoU) Loss. This combination facilitates more stable and accurate localization by improving the regression effectiveness and speeding up the computation.

The specific formulation for Focal Loss is given by:
\begin{equation}
	\text{L}_{\text{MultiFocal}}(p_i,c_i) = -(1-\alpha) \times (1-c_i)^\gamma \times c_i \times \log(p_i)
\end{equation}
Here, \( p_i \) represents the predicted probability of the model for the \( i^{th} \) category, and \( c_i \) is the true label for the \( i^{th} \) category, typically represented as a one-hot encoded vector where the correct category has a value of 1, and all others are 0. The parameter \( \alpha \) serves as a balancing factor to adjust the relative importance of different categories, and \( \gamma \) is a focusing parameter designed to decrease the contribution of easy-to-classify examples and increase the focus on those that are more difficult to classify, thereby improving the robustness and accuracy of the model in distinguishing more subtle or challenging behaviors.

To accurately quantify the discrepancy between the predicted and actual bounding boxes in object detection tasks, we employ the Smooth-L1 loss. This loss function is particularly chosen for its robustness against outliers compared to traditional L2 loss, and its effectiveness in handling errors within certain thresholds. The computation of Smooth-L1 loss is defined as follows:

\begin{equation}
	\text{Smooth}_{L1}(b_i, b_{i}') = 
	\begin{cases}
		0.5 \times (b_{i} - b_{i}')^2, & \text{if } \left| b_{i} - b_{i}' \right| < 1 \\
		\left| b_{i} - b_{i}' \right| - 0.5,   & \text{otherwise}
	\end{cases}
\end{equation}

Here, \( b_{i} \) represents the true bounding box, and \( b_{i}' \) is the predicted bounding box. The function transitions between two behaviors based on the magnitude of the error between the predicted and actual values:
1) For errors less than one unit, the Smooth-L1 loss approximates a quadratic function, where it behaves similarly to L2 loss but with less sensitivity to slight errors. This characteristic is crucial for maintaining stable and efficient gradient descent, as it avoids large updates that could destabilize the learning process.
2) For errors larger than one unit, it behaves like L1 loss, which is linear. This feature prevents the loss from disproportionately escalating due to large errors, often referred to as outliers, thereby ensuring that the model's updates remain significant yet bounded. By incorporating these characteristics, Smooth-L1 loss effectively balances the benefits of L1 and L2 losses, making it an ideal choice for regression problems in object detection where the precision of bounding box predictions is critical.

The Generalized Intersection over Union (GIoU) loss extends the concept of traditional Intersection over Union (IoU) by incorporating a penalty term that accounts for the relationship between the bounding boxes and the smallest enclosing box that contains both. This metric is particularly valuable for object detection tasks as it enhances the accuracy of bounding box predictions. The GIoU is calculated as follows:

\begin{equation}
	\text{GIoU}(b_i, b_{i}') = \frac{|b_i \cap b_{i}'|}{|b_i \cup b_{i}'|} - \frac{|C(b_i, b_{i}') \setminus (b_i \cup b_{i}')|}{|C(b_i, b_{i}')|}
\end{equation}

Here, \( b_i \) and \( b_{i}' \) represent the ground truth and predicted bounding boxes, respectively. \( |b_i \cap b_{i}'| \) denotes the area of their intersection, and \( |b_i \cup b_{i}'| \) denotes the area of their union. The term \( C(b_i, b_{i}') \) refers to the smallest enclosing box that is axis-aligned and contains both \( b_i \) and \( b_{i}' \). The area \( |C(b_i, b_{i}') \setminus (b_i \cup b_{i}')| \) represents the area of the smallest enclosing box that is not covered by the union of the two bounding boxes.

The GIoU loss introduces a penalty proportionate to the excess space within the enclosing box that is not occupied by the bounding boxes, thus encouraging the model to predict bounding boxes that not only overlap greatly with the ground truth but also are tightly bound around it. This added term helps in situations where traditional IoU would return a high score despite poor alignment of the boxes, by considering the overall configuration and encouraging tighter enclosures. By minimizing GIoU loss, the model is trained to enhance both the precision and alignment of the bounding box predictions, making this metric a robust choice for improving the localization accuracy of detected objects.

Finally, the Hungarian Algorithm is employed to establish an exact match between the predicted bounding boxes and the ground truth. This algorithm optimizes the matching to minimize the overall loss by considering the costs associated with each potential pairing, based on classification and bounding box regression losses.

\begin{algorithm}[H]
	\caption{Hungarian Matching Algorithm}
	\begin{algorithmic}
		\STATE 
		\STATE {\textbf{Input:}}
		\STATE Set of predicted bounding boxes for prediction instance $\sigma_i$:\\
		$\hat{y}_{\sigma_{i}} = \{\hat{y}_1, \hat{y}_2, \ldots, \hat{y}_m\}$
		\STATE Set of ground truth bounding boxes:\\
		$y_i = \{y_1, y_2, \ldots, y_n\}$
		\STATE {\textbf{Output:}}
		\STATE Optimal matching of targets and prediction frames:\\		
		$L_{\text{Hungarian}}(y_i, \hat{y}_{\sigma_{i}})$		
		\STATE \textbf{for} each predicted box $\hat{y}_{\sigma_{i}}$ and truth box $y_j$ \textbf{do}
		\STATE \hspace{0.5cm} Compute classification loss: $L_{\text{MultiFocal}}(p_i, c_i)$;
		\STATE \hspace{0.5cm} Compute bounding box loss: \\
		\STATE \hspace{0.5cm} $L_{\text{box}}(b_i, b_{i}^{'}) = \alpha L_{\text{Smooth-L1}}(b_i, b_{i}^{'}) + \beta L_{\text{GIoU}}(b_i, b_{i}^{'})$;		
		\STATE \textbf{end for}
		\STATE Compute total loss: \\
		$L_{\text{Hungarian}} = \sum (L_{\text{cls}}(p_i, c_i) + L_{\text{box}}(b_i, b_{i}^{'}))$		
	\end{algorithmic}
	\label{alg1}
\end{algorithm}

The weights for the three loss components are crucial for model performance optimization. By default, we reference benchmark models to set initial weight values. Typically, the weight ratio between the Intersection over Union loss and the bounding box loss is set to 1:2. Several sets of weight values are compared experimentally to find the most effective distribution for our specific application needs.

%
%
%
%
%
%
%

\section{Experimental Results and Analysis} \label{EXP}

In this section, we demonstrate the effectiveness of the proposed SCB-DETR model by conducting a series of experiments.

\subsection{SCBehavior Dataset}

While a plethora of datasets such as ImageNet, COCO, and VOC has significantly advanced the field of computer vision, particularly in object detection tasks, there remains a distinct lack of specialized resources dedicated to the educational domain. Specifically, there is a noticeable gap in comprehensive datasets aimed at detecting student behavior within classroom settings. To address this deficiency, our research team has developed the SCBehavior dataset, an innovatively designed collection of images meticulously curated to accurately capture and categorize classroom behavior.

\subsubsection{Dataset Construction}
The SCBehavior dataset encompasses 1,346 real-life images, each annotated to reflect a wide array of student behaviors observed in actual classroom environments. To ensure relevance and utility, these images have been classified into seven distinct behavioral categories: writing, reading, looking up, turning head, raising hand, standing, and discussing. This classification not only mirrors the diverse spectrum of student activities but also aligns with the typical pedagogical observations pertinent to educational research.

Given the structured nature of classroom settings, our annotation process was strategically designed to accommodate the uniform seating arrangements commonly found in educational institutions. Consequently, students positioned at the forefront of the classroom were marked as large targets, while those at the periphery were labeled as small targets. This targeted approach allows for nuanced analysis and application of computer vision techniques tailored to varying distances and perspectives within the classroom.

\subsubsection{Dataset Statistics}
To ensure the SCBehavior dataset serves as a robust tool for educational advancements, we have committed to conducting extensive experimental studies to validate its effectiveness and applicability. These studies are intended to rigorously test the dataset's capacity to support the development of automated systems for student behavior analysis, thereby facilitating enhancements in teaching strategies and educational outcomes.

The SCBehavior dataset not only fills a critical void in educational data resources but also sets a precedent for future research in this area. The details and statistics of the SCBehavior dataset are summarized in the Table \ref{tab1}, which showcases the comprehensive nature of the data collected and the meticulous attention to detail employed in its curation.

\begin{table}[h]
\centering
\caption{Statistics of SCBehavior Dataset.} \label{tab1}
\begin{tabular}{cccccc}
\hline
\multicolumn{1}{l}{Number} & Behaviors & Number of Labels & Train  & Validate  & Test \\ \hline
1& Write     & 1025             & 452    & 491  & 82     \\
2& Read      & 1075             & 810    & 139  & 126    \\
3& Lookup    & 5725             & 3620   & 1656 & 449   \\
4& Trun head & 1025             & 748    & 117  & 160   \\
5& Raise hand & 725              & 561    & 82   & 82   \\
6& Stand     & 94               & 50     & 30   & 14   \\
7& Discuss   & 242              & 172    & 50   & 20   \\ \hline
\end{tabular}
\end{table}

\subsection{Experimental Setup}

To conduct our experiments, we have configured a robust hardware and software environment. Our hardware setup includes a server equipped with a 15 vCPU AMD EPYC 7543 32-Core Processor and a single NVIDIA A5000 GPU with 24 GB of VRAM, ensuring sufficient computational power for intensive deep learning tasks. On the software side, our experiments are run using PyTorch version 1.10.0, supplemented by Python 3.8 and CUDA 11.3, to leverage the latest optimizations and support for deep learning frameworks. 

During the training phase, we employ the AdamW optimizer, a variant of the Adam optimization algorithm that includes decoupled weight decay. This choice is informed by AdamW's proven effectiveness in handling sparse gradients and preventing overfitting in complex models. The training is configured with a batch size of 2, a learning rate of 0.0002, and a weight decay factor of 0.0001. These settings are meticulously chosen to balance the training speed and convergence stability, optimizing the adjustment of model parameters.

Furthermore, the SCB-DETR framework, which is central to our study, utilizes pre-trained weights from the Deformable-DETR model to enhance its feature extraction capabilities. Training the SCB-DETR on the SCBehavior dataset ensures that the model can effectively leverage its architecture's strengths while being finely tuned to the specific task of recognizing student behavior in classroom settings. Consistency in hyperparameters across training sessions is meticulously maintained to ensure reproducibility and reliability of the results.


\subsection{Evaluation Metrics}

To ensure a thorough evaluation of our model's performance in detecting student behavior in classroom settings, we employ a range of established metrics. These include mAP, Average Precision (AP), Accuracy, and Recall. Each metric provides insights into different aspects of model performance, helping to assess both the effectiveness and reliability of our approach.

\subsubsection{Precision and Recall}
Precision and recall are fundamental to understanding the model's performance at identifying relevant instances among the retrieved cases. They are defined as follows:

\begin{equation}
	\text{Precision} = \frac{TP}{TP + FP}
\end{equation}

\begin{equation}
	\text{Recall} = \frac{TP}{TP + FN}
\end{equation}

Where \(TP\) (True Positives) represents correctly identified positive cases, \(FP\) (False Positives) represents negative cases incorrectly identified as positive, and \(FN\) (False Negatives) represents positive cases that were not identified. These metrics are crucial for computing further evaluation metrics like AP and mAP.

\subsubsection{Average Precision (AP) and Mean Average Precision (mAP)}
AP provides a single-figure measure of quality across recall levels, essentially giving us the average of precision scores calculated at each threshold. It is particularly useful when precision varies with recall. The AP for a single category is computed as the area under the Precision-Recall (P-R) curve, derived by plotting recall on the x-axis against precision on the y-axis:

\begin{equation}
	AP = \int_{0}^{1} Precision \, d(Recall)
\end{equation}

The mAP is the mean of APs across all categories, providing a single measure to evaluate the overall performance of the detection system across different object categories:

\begin{equation}
	mAP = \frac{\sum_{j=0}^n AP(j)}{n}
\end{equation}

Here, \(n\) represents the number of categories. The mAP is a comprehensive metric that averages the AP obtained for each class, reflecting the model’s ability to detect each class with both precision and recall taken into account.

These metrics collectively enable a robust assessment of our model, highlighting its capabilities and areas for improvement in the context of automated classroom behavior analysis.

\subsection{Baselines}

To demonstrate the effectiveness of our model, we conducted a comparative analysis against several prominent baseline models from both academia and industry. This comparison allows for a comprehensive assessment of our model’s performance across diverse environments and tasks, confirming its utility and reliability for student behavior recognition applications.

\subsubsection{YOLOv5 \cite{bib25}}
YOLOv5 is renowned for its real-time object detection capabilities. This model is designed to perform detection and classification simultaneously, enhancing efficiency. YOLOv5 introduces improvements over its predecessors in accuracy, speed, and modularity, employing CSP-Darknet53 as its backbone with Cross Stage Partial (CSP) connections to optimize gradient flow and computational efficiency. Additionally, it utilizes a Path Aggregation Network (PANet) to enhance feature integration across scales.

\subsubsection{YOLOv7 \cite{bib15}}
YOLOv7 extends the YOLO family with several architectural enhancements that boost performance, striking a balance between speed and accuracy. This model incorporates Efficient Layer Aggregation Networks (ELAN) and RepVGG blocks, which simplify the network structure while improving feature extraction and reducing memory usage. For our comparative analysis, we chose the YOLOv7-tiny variant due to its parameter efficiency, making it comparable to our model in terms of resource utilization.

\subsubsection{SSD \cite{bib16}}
The Single Shot MultiBox Detector (SSD) is another efficient model known for its balance between detection speed and accuracy. Utilizing a one-shot detection mechanism and multi-scale feature maps, SSD is well-suited for real-time applications. Although it sometimes struggles with very small or densely packed objects, SSD's capabilities make it an important reference model in our comparison.

\subsubsection{Faster R-CNN \cite{bib14}}
Faster R-CNN employs a two-stage process for high accuracy detection. The first stage generates region proposals through a Region Proposal Network (RPN) \cite{Tang2018}, and the second stage classifies and refines these proposals. Given its proven effectiveness in complex scene analysis, Faster R-CNN is included as a baseline model to evaluate its performance in accurately detecting and localizing student behaviors in classroom settings.

\subsubsection{Mask DINO \cite{Li2023l}}
Mask DINO is an extension of DINO \cite{bib10}. This enhancement leverages the query embeddings generated by DINO, employing a dot product with a high-resolution pixel embedding map to predict a set of binary masks for each image segment. This adaptability makes Mask DINO particularly valuable in scenarios where both precise object detection and detailed segmentation are required, providing significant benefits from the synergy of combined detection and segmentation tasks.

\subsubsection{Deformable-DETR \cite{Wang2023g}}
Deformable-DETR enhances the original DETR framework by incorporating deformable attention mechanisms, which improve the handling of small objects and accelerate convergence. This model’s ability to efficiently extract features and adapt to dynamic scenarios makes it an ideal baseline for assessing our model's performance in recognizing diverse and detailed student behaviors.

By comparing our proposed model with these established detection frameworks, we aim to highlight its strengths and potential improvements, ensuring it meets the rigorous demands of real-world educational applications.

\subsection{Comparison with Baseline Methods}

This section details the SCBehavior dataset, our experimental setup, and the results obtained, which are presented to ensure the reproducibility of our findings. Our study benchmarks the proposed model against several prominent target detection algorithms. To showcase the enhancements our model brings, we performed a series of experiments using the SCBehavior dataset and assessed the outcomes against various established detectors. We compared our approach with leading methods including YOLOv5\cite{bib25}, the most recognized single-stage object detection model, as well as YOLOv7\cite{bib15}, SSD\cite{bib16}, and Faster R-CNN\cite{bib14}. Additionally, our comparison extends to refined DETR models such as DINO \cite{Li2023l} and Deformable-DETR \cite{Wang2023g}, which are advancements in the DETR architecture.

\begin{table}[h] 
	\centering 
	\caption{Performance comparison of experimental results with baseline methods.}	\label{tab2}
	\begin{tabularx}{0.51\textwidth}{cccccc} 
		\hline
		\textbf{Methods}   &\textbf{Models}   & \textbf{mAP} & \textbf{AP50} & \textbf{Recall} & \textbf{Parameter} \\ \hline
		Liu et al.\cite{bib16} & SSD             & 0.501        & 0.736         & 0.733           & \textbf{24.5M} \\
		Ren et al.\cite{bib14}&Faster-RCNN     & 0.576        & 0.804         & 0.687           & 41.7M \\
		Jocher et al.\cite{bib25}&YOLOv5-L     	  & 0.547        & 0.816         & 0.724           & 46.5M \\
		Jocher et al.\cite{bib25}&YOLOv5-X        & 0.573        & 0.835         & 0.709           & 86.7M \\
		Wang et al.\cite{bib15}&YOLOv7       & 0.611        & 0.817         & \textbf{0.785}  & 36.9M \\
		Wang et al.\cite{bib15}&YOLOv7-X       & 0.579        & 0.797         & 0.768           & 71.3M \\
		Li et al.\cite{Li2023l}&Mask DINO           & 0.564        & 0.766         & 0.782           & 47.5M \\
		Wang et al.\cite{Wang2023g}&De-DETR & 0.544        & 0.654         & 0.722           & 41.1M \\ \hline
		&\textbf{Ours} & \textbf{0.626} & \textbf{0.877} & 0.757 & 43.1M \\
		\hline
	\end{tabularx}
\end{table} 

All experiments were conducted using consistent training and test datasets, and followed identical data augmentation strategies to ensure fairness in comparison. For evaluation metrics, we employed AP, mAP, and Recall, following the COCO evaluation standards. Our proposed model was compared against eight established object detection models, including variations of YOLOv5 and YOLOv7, as well as SSD, Faster R-CNN, DINO, and Deformable-DETR. The performance metrics were carefully analyzed, particularly focusing on the mAP to gauge the overall effectiveness of each model.

As shown in Table \ref{tab2}, it clearly demonstrate that our model surpasses the baseline models in terms of mAP, achieving a peak of 62.6\%, which is 1.5\% higher than the nearest competitor, YOLOv7, and 8.2\% higher than the baseline Deformable-DETR model.

\begin{figure*}[ht]
	\centering
	\subfigure[$mAP$]{\includegraphics[width=0.49\linewidth]{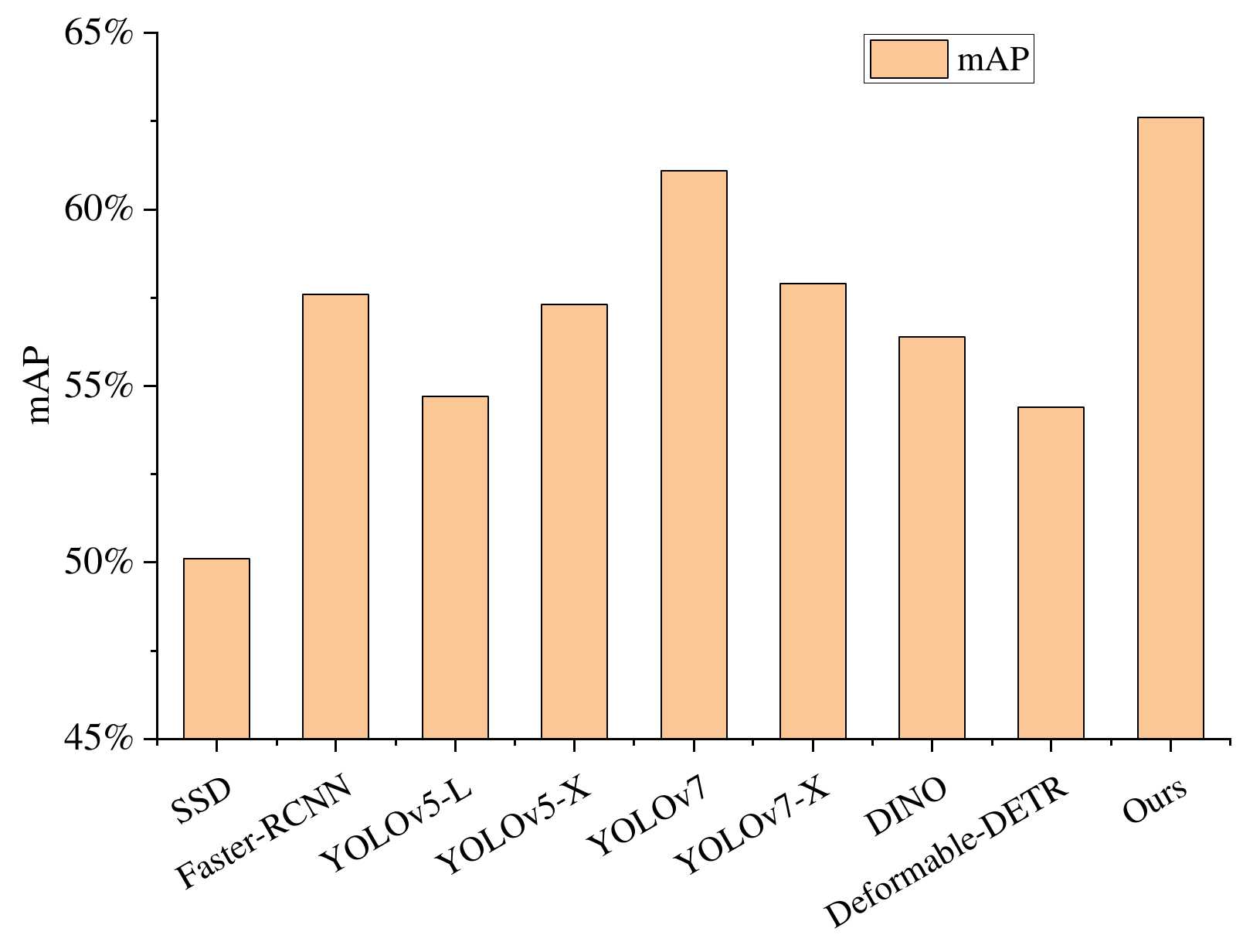}}
	\subfigure[$AP50$]{\includegraphics[width=0.49\linewidth]{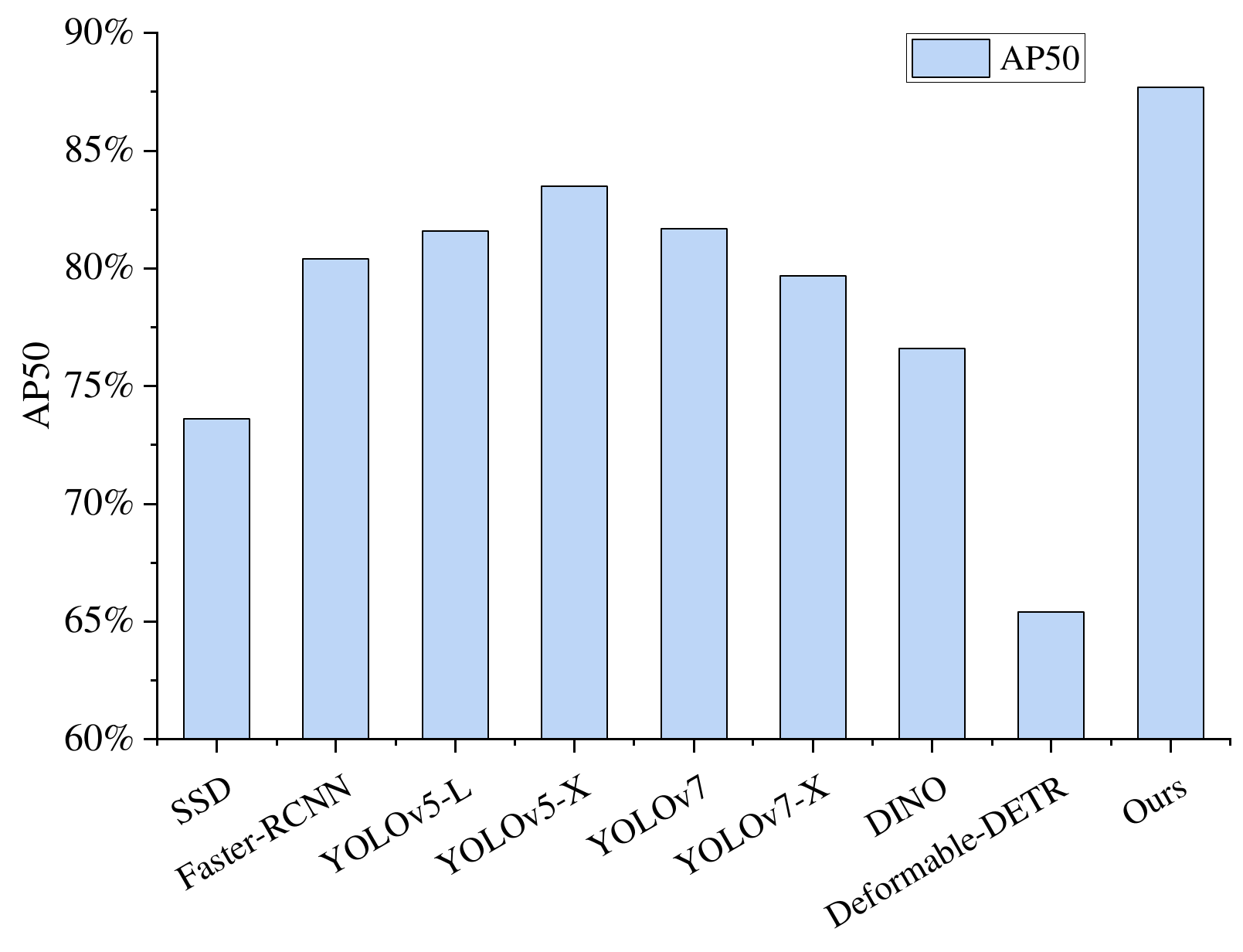}}
	\subfigure[$Recall$]{\includegraphics[width=0.49\linewidth]{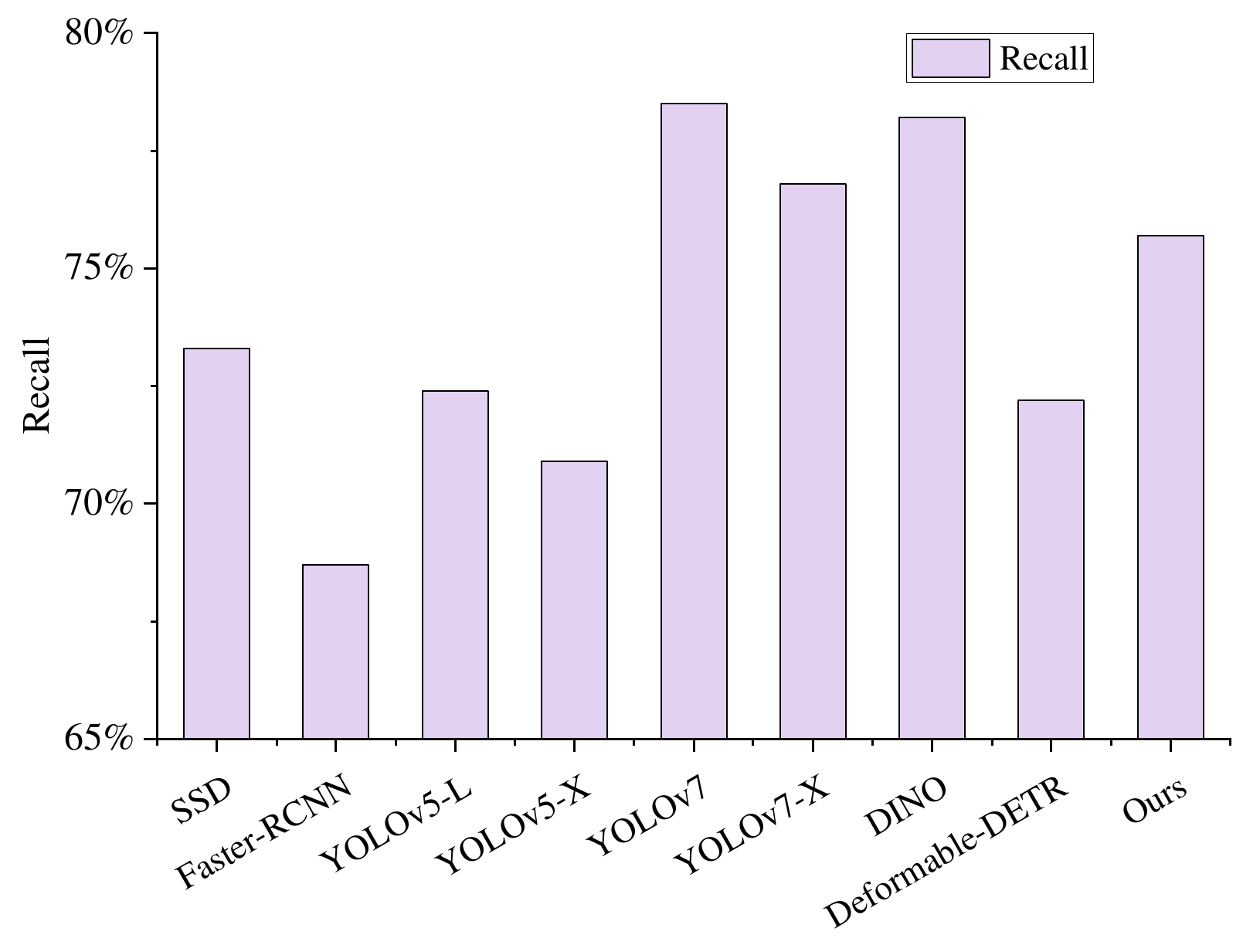}}
	\subfigure[$mAP$ and parameter numbers]{\includegraphics[width=0.49\linewidth]{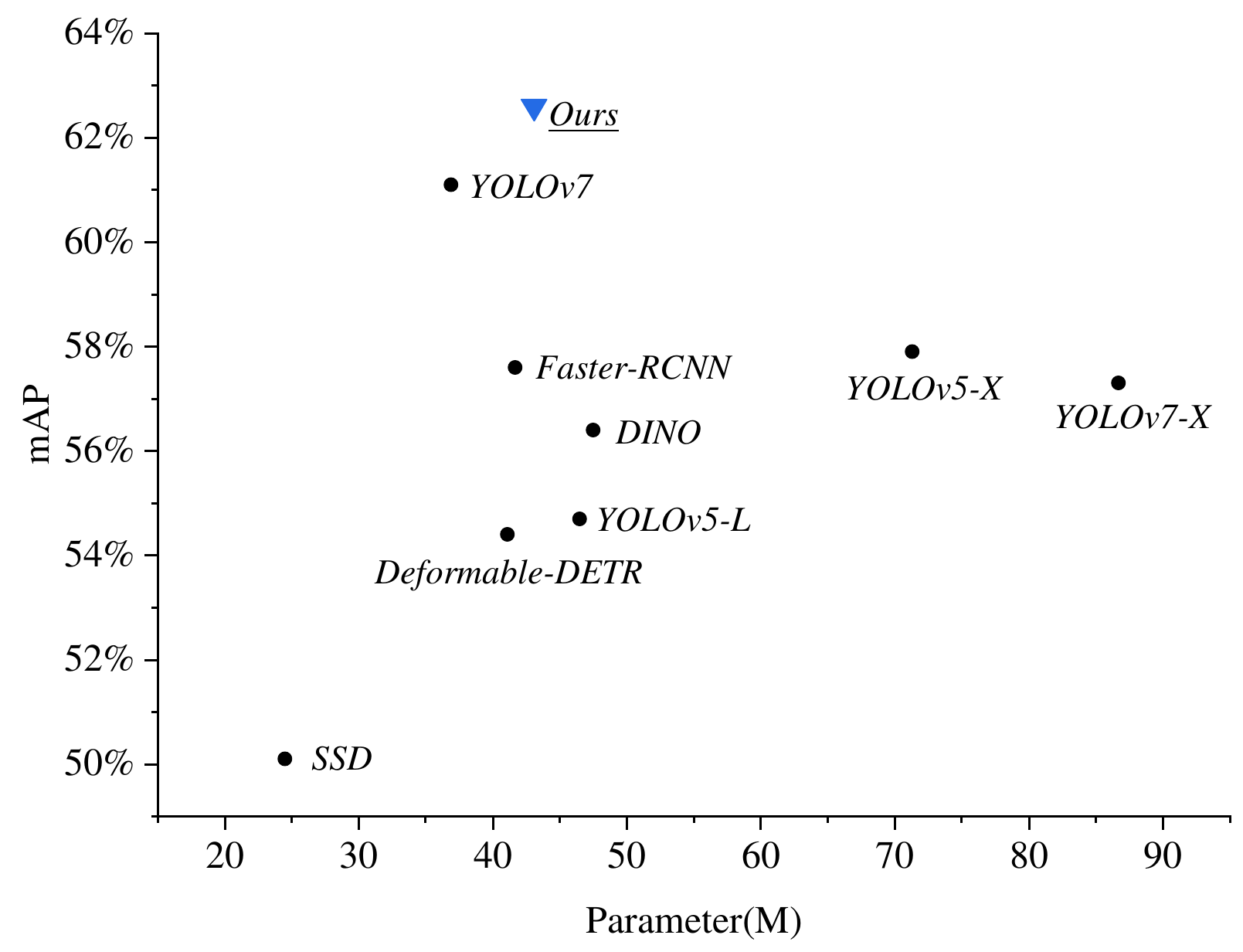}}	
	\caption{Comparison with baseline methods. (a) is the $mAP$ of the baseline methods and the proposed method. (b) is the $AP50$ of the baseline methods and the proposed method. (c) is the $Recall$ of the baseline methods and the proposed method. (d)is the parameter numbers of the baseline methods and the proposed method.}\label{fig7}
\end{figure*} 

Additionally, as shown in Fig. \ref{fig7}, our model achieved the highest AP50 score among all compared models, further establishing its superiority in detecting and classifying behaviors accurately within classroom settings. At the same time, as shown in Fig. \ref{fig7}, our model achieves a balance between performance and the number of model parameters.

The empirical evidence underscores the advanced detection capabilities of our model, showcasing its suitability for real-world educational applications. By delivering superior accuracy and efficiency, our model holds promise for enhancing automated analysis and monitoring of student behavior, ultimately contributing to better educational outcomes.

\subsection{Ablation Experiments}

To evaluate the impact of the LKNeXt backbone and the HFFA module on enhancing model performance, we conducted ablation studies using Deformable-DETR as the baseline model. Initially, we replaced the baseline's ResNet backbone with ConvNeXt, noted for its greater efficiency. Subsequently, we upgraded ConvNeXt to the more robust LKNeXt. We then integrated both the LKNeXt backbone and the HFFA module into the network structure, testing their performance on our proprietary SCBehavior dataset. The experimental results are detailed in Table \ref{tab3}.

\begin{table}[h]
	\centering
	\caption{Experimental effects of ablation of the proposed model.} \label{tab3}
	\begin{tabular}{ccccc}
		\hline
		Groups & Models & mAP       &   AP50    &   Recall \\ \hline
		1 & Base (Deformable-DETR)                     & 0.544     &   0.654   &   0.722    \\
		2 & Base + LKNeXt              & 0.556     &   0.723   &   0.741    \\
		3 & Base + ConvNeXt+HFFA       & 0.612     &   0.845   &   0.761    \\
		\hline
		Ours & Base + LKNeXt + HFFA    & \textbf{0.626$\uparrow$}     &   \textbf{0.877$\uparrow$}   &   \textbf{0.757$\uparrow$}     \\
		\hline
	\end{tabular}
\end{table}

\subsubsection{Effectiveness of LKNeXt} 

\begin{figure}[h]
	\centering
	\includegraphics[width=0.49\textwidth]{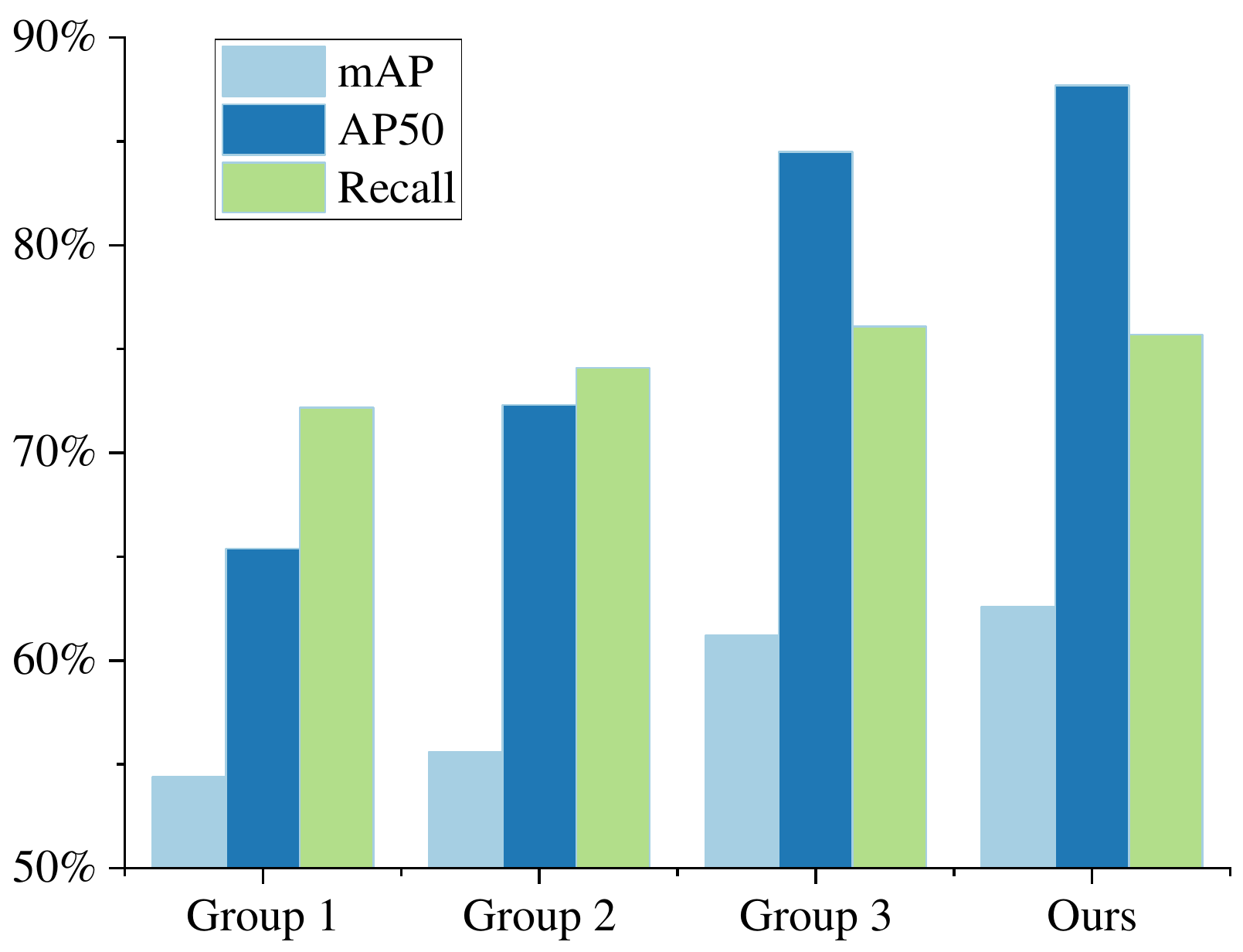}
	\caption{Ablation experiments with different module combinations.}
	\label{fig8}
\end{figure}

The incorporation of the LKNeXt backbone led to noticeable improvements in model performance. Specifically, transitioning from the base Deformable-DETR model to the LKNeXt enhanced configuration resulted in a significant increase in mAP and Recall. Furthermore, as shown in Fig. \ref{fig8}, the addition of LKNeXt contributed to a nearly 1\% increase in mAP over the ConvNeXt backbone configuration, emphasizing the backbone's superior capability to capture detailed features necessary for precise behavior detection in educational settings.

These results highlight the advantages of LKNeXt in improving the overall detection accuracy and efficiency of the model, validating its effectiveness in complex object detection scenarios such as those encountered in classroom behavior analysis.

\subsubsection{Effectiveness of HFFA}

The addition of the HFFA module significantly enhances the performance of our baseline model. By incorporating HFFA into the architecture, we observed substantial improvements in detection metrics. Specifically, as shown in Fig. \ref{fig8}, the introduction of HFFA resulted in an increase in the mAP by 5.6\%, demonstrating a marked enhancement from the baseline configuration. Additionally, the Recall metric showed a noteworthy increase of 2\%, highlighting the module's effectiveness in improving the model’s ability to correctly identify relevant instances.

These results affirm the HFFA module's capacity to augment the detection capabilities of the system, facilitating more accurate and reliable performance across varied detection scenarios. The improvements in mAP and Recall are indicative of the module's robustness, particularly in complex environments where precision and recall are critical for successful detection.

\subsubsection{Training Performance in Different Experimental Groups}

\begin{figure}[h]
	\centering
	\includegraphics[width=0.5\textwidth]{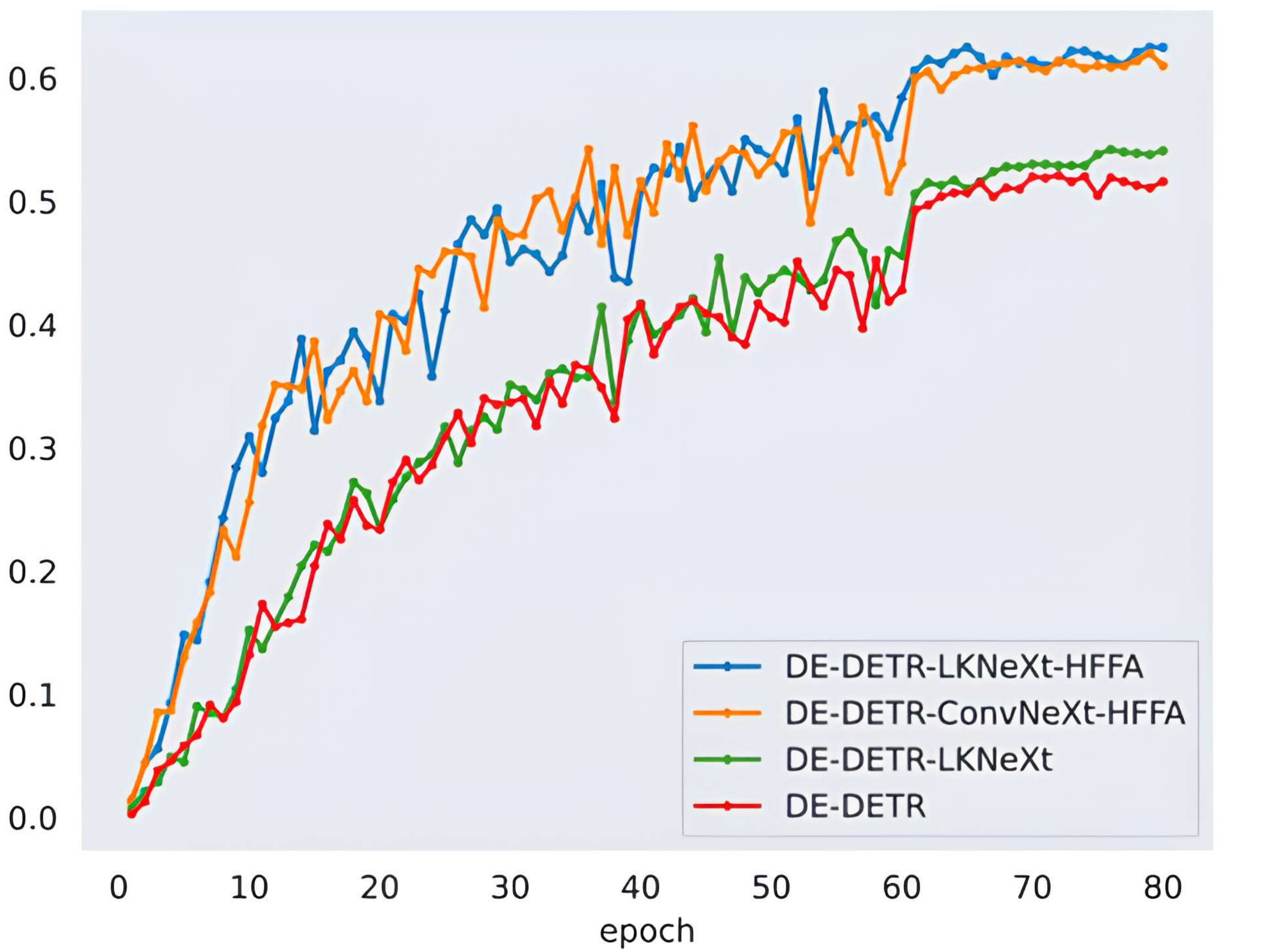}
	\caption{Comparison of training performance across different experimental groups.}
	\label{fig9}
\end{figure}

Fig. \ref{fig9} illustrates the mAP (0.5 : 0.95) progression over various training cycles. The data reveals consistent performance enhancement across all models with extended training periods. Notably, the DE-DETR-LKNeXt+HFFA configuration exhibits superior performance throughout the training cycles, attributed to the LKNeXt backbone's optimized feature extraction capabilities and the enhanced feature fusion provided by the HFFA module.

While the DE-DETR-ConvNeXt-HFFA configuration slightly lags behind the DE-DETR-LKNeXt+HFFA, it still outperforms the baseline DETR model, which incorporates only the LKNeXt backbone without further enhancements. Additionally, the DE-DETR-LKNeXt setup, which integrates the LSK block, also surpasses the original DETR model, further emphasizing the beneficial impact of backbone enhancements on model performance.

Interestingly, as training progresses, the performance gap between DE-DETR-LKNeXt+HFFA and DE-DETR-ConvNeXt-HFFA configurations narrows. This observation suggests that the incremental benefits of the HFFA module may diminish over extended training periods. Nonetheless, our results clearly demonstrate that strategic architectural enhancements, particularly those improving feature extraction and fusion, can significantly elevate the target detection capabilities of DETR models. These findings not only underscore the potential of architectural innovations in enhancing DETR models but also serve as a valuable reference for ongoing and future developments in DETR architecture. This insight is instrumental for researchers in the field of computer vision seeking to refine target detection performance.

\subsection{Case Study}

In this section, we evaluate the effectiveness of the proposed SCB-DETR by examining various case studies.

\subsubsection{Detection Performance in Real Classroom Scenarios}

\begin{figure*}[h]
	\centering
	\subfigure[Example of a single detection.]{\includegraphics[width=\linewidth]{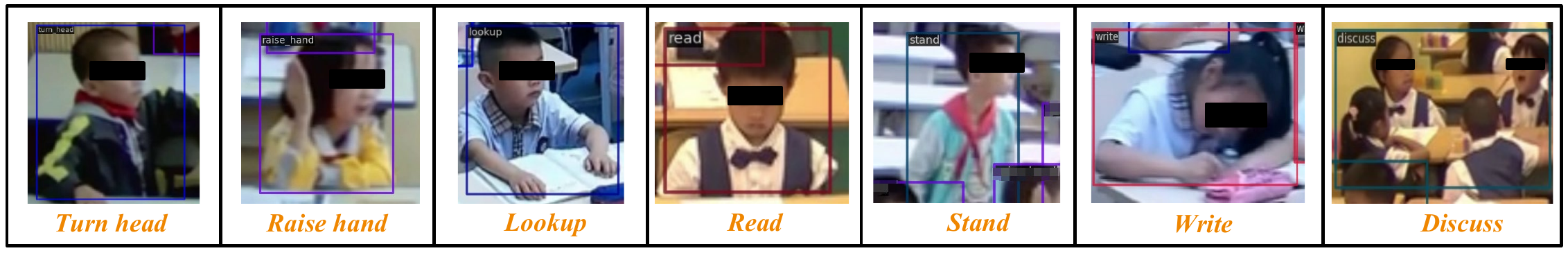}}
	\subfigure[Examples of multiple detections in a real classroom.]{\includegraphics[width=\linewidth]{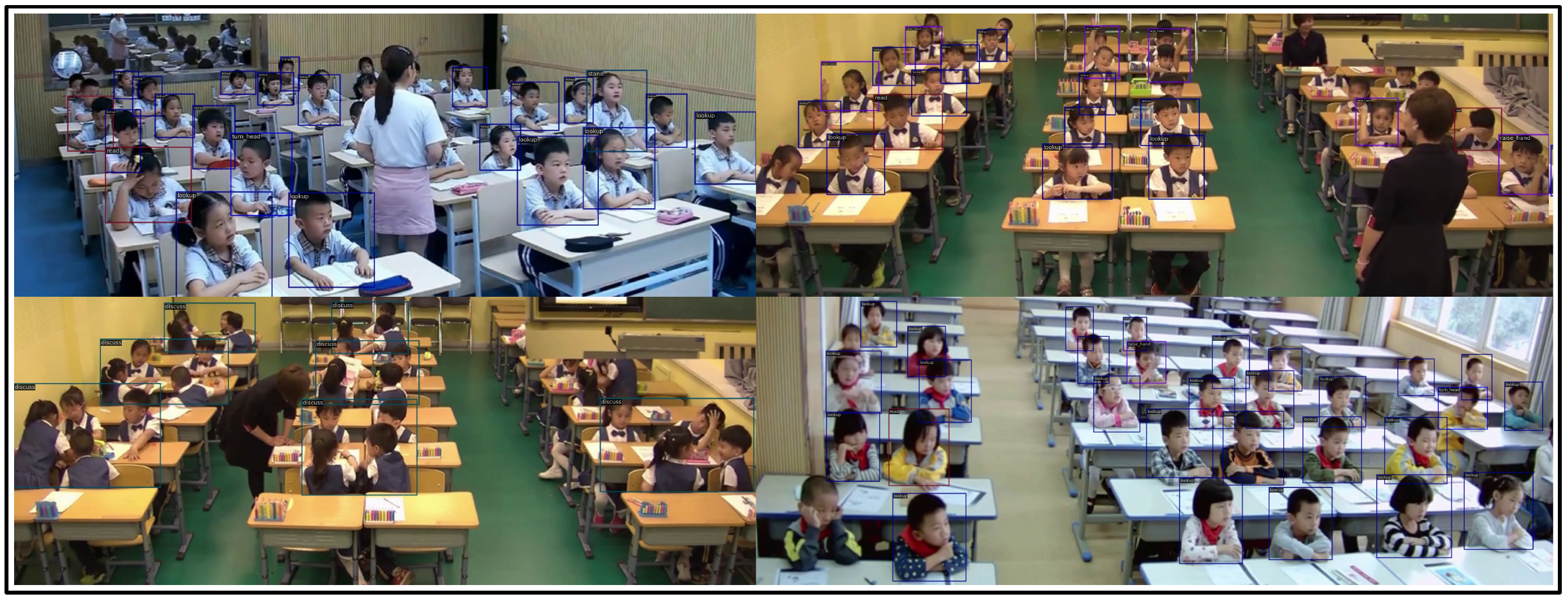}}
	\caption{Detection results using the proposed SCB-DETR model.}
	\label{fig10}
\end{figure*}

The effectiveness of the SCB-DETR model in real classroom settings is illustrated in Fig. \ref{fig10}, which showcases the model's capability to accurately detect student behaviors across various classroom environments. As demonstrated, the model successfully identifies individual and group activities within the dynamic context of classroom interactions, underscoring its robustness and reliability.

This performance not only validates the model's practical application in traditional classroom settings but also highlights its potential for broader implementation in cloud-based classrooms and online education platforms. By effectively recognizing diverse student behaviors, the SCB-DETR model provides a solid foundation for enhancing educational engagement and interaction in both physical and virtual learning environments.

\subsubsection{Confusion Matrix}

\begin{figure}[h]
	\centering
	\includegraphics[width=0.49\textwidth]{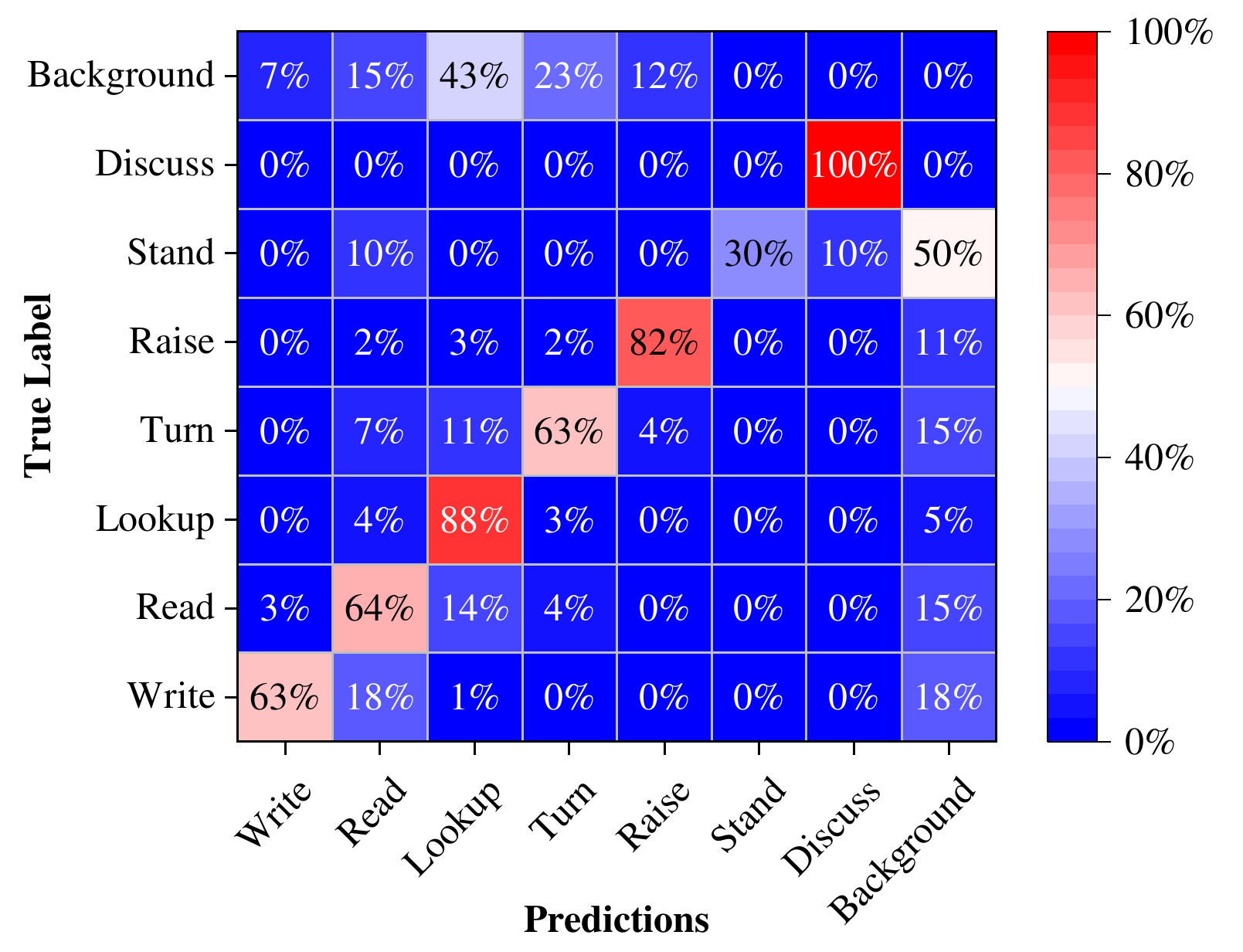}
	\caption{Confusion matrix illustrating detection accuracies for different student behaviors.}
	\label{fig11}
\end{figure}

To evaluate the accuracy and effectiveness of our model in detecting various student behaviors in classroom settings, we utilized a confusion matrix, as depicted in Fig. \ref{fig11}. This matrix elucidates the model’s performance across seven distinct classroom behaviors, providing a clear visual representation of where the model excels or underperforms.

The confusion matrix reveals that our model exhibits high accuracy in detecting behaviors that involve significant body movements, such as raising hands, turning heads, looking up, and engaging in discussions. These activities are generally more distinguishable due to their dynamic nature.

Conversely, the model demonstrates lower accuracy in identifying behaviors such as reading and writing. These activities often involve subtler body movements and are less distinct from one another, posing challenges for the model due to their similar postures and minimal interaction with the surrounding environment.

Overall, the analysis highlights the strengths and weaknesses of our model in real-world classroom applications. While the model shows commendable generalization capabilities and effectively leverages a large convolutional kernel to detect a broad spectrum of behaviors, it still faces challenges with behaviors that are visually similar and less contextually varied. This insight into the model’s performance underscores the need for further refinement, particularly in enhancing its sensitivity to subtle behaviors.

\subsubsection{Precision-Recall Curve}

\begin{figure}[t]
	\centering
	\includegraphics[width=0.49\textwidth]{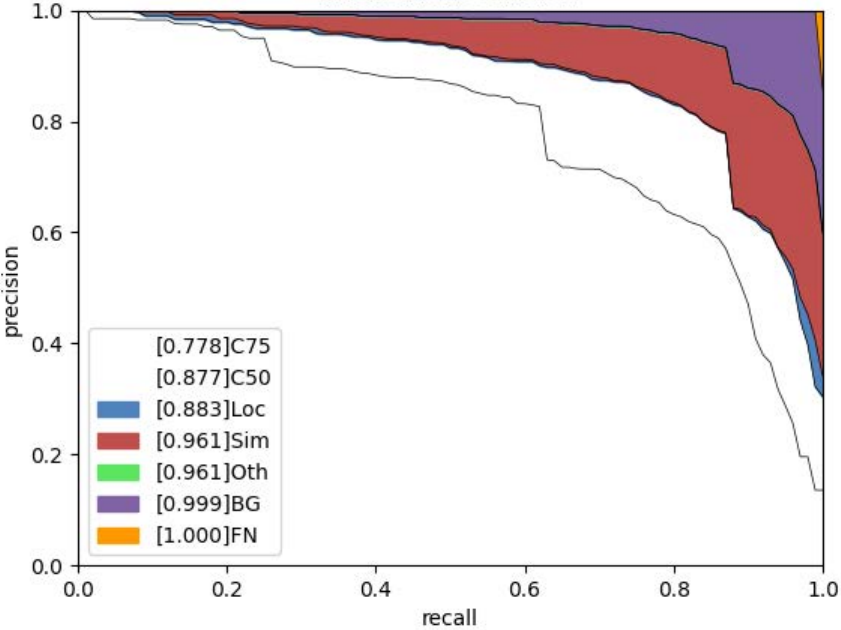}
	\caption{Precision-Recall (P-R) Curve for SCB-DETR.}
	\label{fig12}
\end{figure}

The P-R curve depicted in Fig. \ref{fig12} evaluates the performance of our SCB-DETR model across various categories and sizes of student behavior, utilizing distinct color-coded curves for different evaluation metrics or detection conditions. The interpretation of the curves are as follows:
\begin{itemize}
	\item The \textbf{C75 and C50 curves} illustrate the model's performance at intersection-over-union (IoU) thresholds of 0.75 and 0.50, respectively. With area under the curve (AUC) values of 0.778 for C75 and 0.877 for C50, these results indicate a decline in performance at higher IoU thresholds, reflecting the increasing challenge of precise detection.
	\item The \textbf{Loc curve}, representing localization accuracy, shows an AUC of 0.883, confirming the model's effectiveness in correctly positioning detected objects within the classroom.
	\item The \textbf{Sim and Oth curves} measure the model's efficacy in distinguishing between similar objects and handling other types of errors, both achieving an AUC of 0.961, demonstrating robust discriminative capabilities.
	\item The \textbf{BG curve} addresses the model's performance in avoiding false positives from the background, with an impressive AUC of 0.999, indicating almost perfect background classification.
	\item The \textbf{FN curve} reflects the absence of false negatives, achieving an AUC of 1.000, suggesting that no positive examples were missed during testing.
\end{itemize}

The P-R curve generally exhibits a decreasing trend in precision as recall increases, a common phenomenon as the model attempts to capture more positive instances, inevitably raising the number of false positives. Nevertheless, the model excels under all test conditions, particularly in minimizing background misclassifications and omissions.

This analysis confirms the robustness and high accuracy of our SCB-DETR model in detecting diverse student behaviors in classroom settings, even at stringent IoU thresholds. Future research will aim to enhance model sensitivity at lower IoU thresholds and reduce the confusion between similar objects, further refining our detection framework.

\subsubsection{Different Loss Settings}

\begin{figure}[t]
	\centering
	\includegraphics[width=0.49\textwidth]{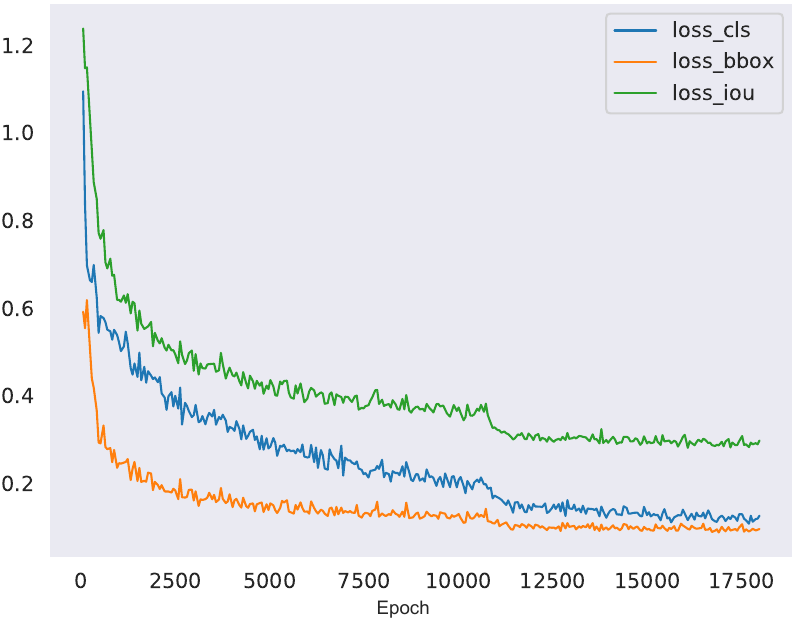}
	\caption{The effect of three loss functions.}
	\label{fig13}
\end{figure}

Fig. \ref{fig13} illustrates the trajectories of three critical loss functions used in training our model: classification loss (loss/cls), bounding box regression loss (loss/bbox), and intersection over union loss (loss/iou). Each curve represents the loss's progression through the training iterations. The analysis are as follows:

\begin{itemize}
	\item The \textbf{classification loss} (depicted by the blue curve) initiates at a high value and exhibits a rapid decline in the initial phases of training, which then transitions into a gradual stabilization. This pattern suggests that the model's capability to correctly classify student behaviors in classroom settings is improving significantly as training progresses.
	\item The \textbf{bounding box regression loss} and \textbf{intersection over union loss} (orange and green curves, respectively) also begin at higher values and follow a similar downward trend, indicating an enhancement in the model’s accuracy for predicting the precise locations of bounding boxes associated with student behaviors.
\end{itemize}

The initial steep descent in all three curves reflects rapid learning and feature extraction from the data, which is typical in the early stages of model training. As training continues, the rate of loss reduction slows, indicating that the model is refining its parameters and approaching a state of convergence—a common phenomenon in neural network training.

Overall, these loss curves demonstrate the continuous improvement of the model across classification and localization tasks during training. The consistent convergence of the loss values points to the stability of the model and the adequacy of the training regimen. This insight into the training dynamics is crucial for understanding the development of the model's performance, particularly its ability to learn and adapt to the complexities of student behavior detection in classroom environments.

\subsection{Sensitivity Analysis}

\begin{figure}[h]
	\centering
	\includegraphics[width=0.49\textwidth]{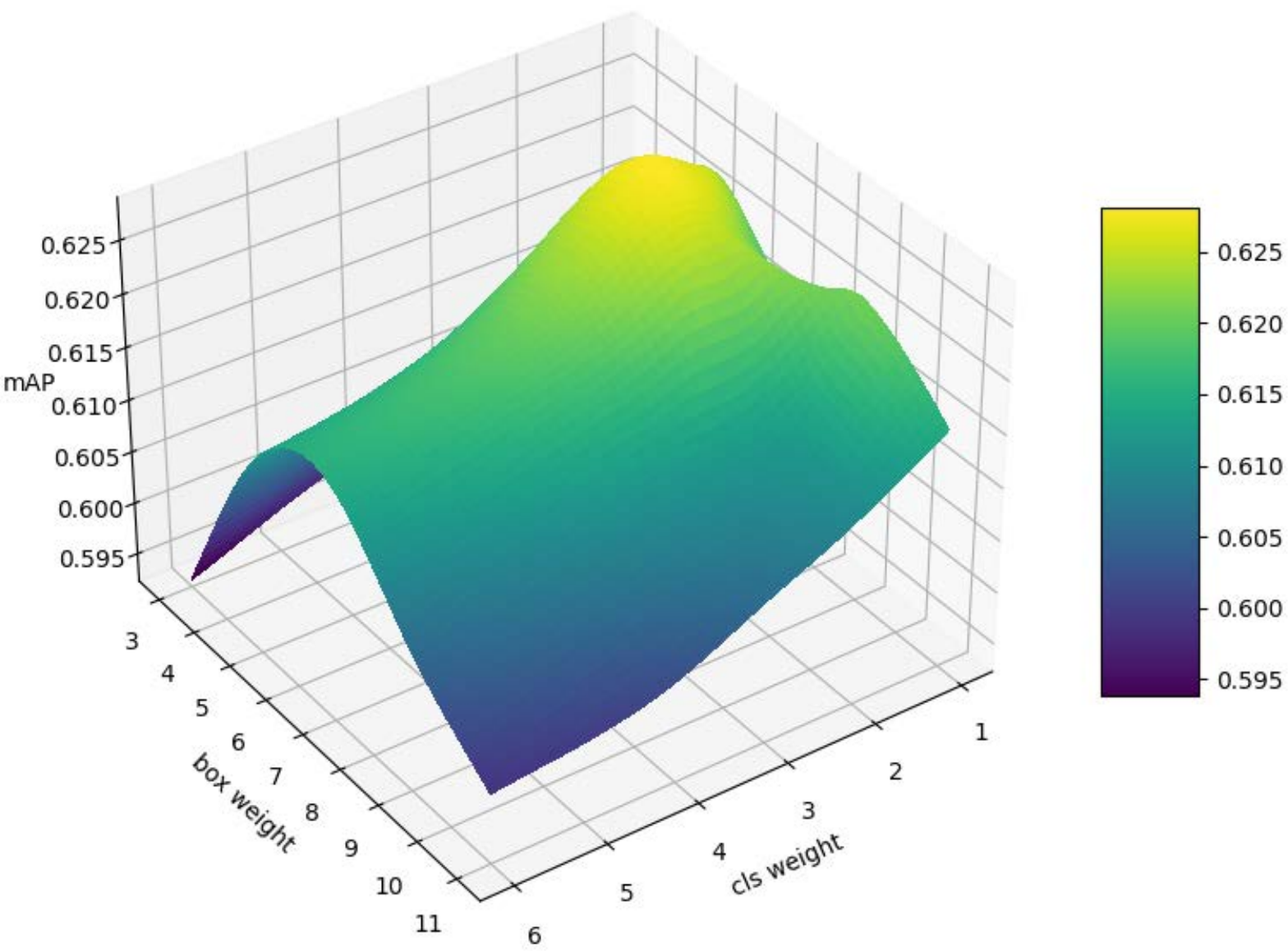}
	\caption{Sensitivity analysis on two hyperparameters.}
	\label{fig14}
\end{figure}

In our pursuit to fine-tune the SCB-DETR model for optimal performance, we conducted a sensitivity analysis of the model's critical hyperparameters, specifically focusing on the Classification Weight (CLS Weight) and the Bounding Box Weight (Box Weight). We varied the Box Weight from 3 to 11 and the CLS Weight from 1 to 6 to evaluate their impact on the model’s effectiveness.

As illustrated in Fig. \ref{fig14}, the optimal performance was observed when the Box Weight was set at 4 and the CLS Weight at 2, achieving the highest mAP of 0.626. This particular combination of weights proved most effective in balancing the trade-offs between classification accuracy and bounding box precision. Given these findings, we have established this weight configuration as the standard setting for our model, ensuring it operates under the most effective parameter settings for detecting student behaviors in classroom environments.

\section{Conclusion} \label{CON}

This study introduces SCB-DETR, an innovative end-to-end network architecture designed to enhance both the effectiveness of detecting students' classroom behaviors and the computational efficiency of the model. Built upon the foundational principles of Deformable DETR, SCB-DETR incorporates the LKNeXt backbone network, which leverages a large convolutional kernel inspired by ConvNeXt, demonstrating superior performance in recognizing student behaviors.

The core innovation of LKNeXt resides in its Large Kernel design, which significantly expands the receptive field, capturing crucial spatial relationships and subtle movements that are pivotal for accurate behavior analysis. To further refine the model's capabilities, upstream features extracted from LKNeXt are processed through our newly developed HFFA. This architecture integrates a multi-scale feature pyramid with a deformable encoder tailored specifically for the nuances of student behavior, replacing traditional encoder layers with a more dynamic structure that enhances feature fusion across various scales.

HFFA’s effectiveness is exemplified by its ability to amalgamate multi-scale information, enriching the detail captured in behavioral features. This is accomplished through a synergistic combination of the NASFCOS-FPN module and a deformable convolutional encoder, complemented by a coordinate attention mechanism. Such integration allows the model to simultaneously consider channel and spatial information, drastically improving the precision of occlusion behavior recognition.

The union of LKNeXt and HFFA signifies a significant advancement in detecting and analyzing student behavior within classroom settings. These enhancements not only boost the model’s accuracy and computational efficiency but also improve its adaptability to diverse educational environments. This progress is crucial for the deployment of smart classroom solutions and cloud-based educational platforms, offering substantial benefits to the field of educational technology.

Looking forward, the development of SCB-DETR sets a robust foundation for future research aimed at refining detection algorithms to handle lower visibility behaviors and minimize confusions in behaviorally similar scenarios. The insights gained from this study will guide the enhancement of smart educational tools, ultimately contributing to more personalized and effective learning experiences.

\bibliographystyle{unsrt}
\bibliography{ref.bib,240826IEEEM.bib,refO.bib,Citations.bib}
\end{document}